\documentclass[10pt,journal,compsoc]{IEEEtran}
\newif\ifpeerreview

 \peerreviewfalse
\usepackage[nocompress]{cite}
\usepackage{url}
\usepackage{color}
\usepackage{amsmath,amssymb,graphicx}

\usepackage{enumitem}
\usepackage{hyperref}

\usepackage{booktabs}
\newcommand{\ra}[1]{\renewcommand{\arraystretch}{#1}}

\usepackage[switch]{lineno}

\newcommand{\paperID}{0003}

\title{Computational Imaging for Long-Term Prediction of Solar Irradiance}

\author{Leron~K.~Julian, %~\IEEEmembership{Student Member,~IEEE,}
	Haejoon Lee,
         Soummya~Kar,
        and~Aswin~C.~Sankaranarayanan%,~\IEEEmembership{Life~Fellow,~IEEE}% <-this % stops a space
\IEEEcompsocitemizethanks{\IEEEcompsocthanksitem All authors are with  the Department
of Electrical and Computer Engineering, Carnegie Mellon University, Pittsburgh, PA 15213.\protect\\
E-mail: \{ljulian,haejoonl,soummyak,saswin\}@andrew.cmu.edu.}% <-this % stops an unwanted space
}

\begin{document}

\IEEEtitleabstractindextext{%
\begin{abstract}
The occlusion of the sun by clouds is one of the primary sources of uncertainties in solar power generation, and is a factor that affects the wide-spread use of solar power as a primary energy source. 
%
%Cloud cover is the largest source of uncertainty due to its intermittent shape and trajectory. 
%
Real-time forecasting of cloud movement and, as a result, solar irradiance is necessary to schedule and allocate energy across grid-connected photovoltaic systems.
Previous works monitored cloud movement using wide-angle field of view imagery of the sky.
However, such images have poor resolution for clouds that appear near the horizon, which reduces their effectiveness for long term prediction of solar occlusion.
Specifically, to be able to predict occlusion of the sun over long time periods,  clouds that are near the horizon need to be detected, and their velocities estimated precisely.
To enable such a system, we design and deploy a catadioptric system that delivers wide-angle imagery with uniform spatial resolution of the sky over its field of view.
To enable prediction over a longer time horizon, we design an algorithm that uses carefully selected spatio-temporal slices of the imagery using estimated wind direction and velocity as inputs.
Using ray-tracing simulations as well as a real testbed deployed outdoors, we show that the system is capable of predicting solar occlusion as well as irradiance for tens of minutes in the future, which is an order of magnitude improvement over prior work.
%
%Our main contribution is to show that by using this system, cloud motion appears true, immaterial of location in the sky which improves long-term prediction of cloud evolution and solar irradiance far above previous works.
\end{abstract}

}

% Make Title
\ifpeerreview
\linenumbers \linenumbersep 15pt\relax 
\author{Paper ID \paperID\IEEEcompsocitemizethanks{\IEEEcompsocthanksitem This paper is under review for ICCP 2024 and the PAMI special issue on computational photography. Do not distribute.}}
\markboth{Anonymous ICCP 2024 submission ID \paperID}%
{}
\fi
\maketitle

% The first section title should be wrapped inside a \IEEEraisesectionheading as follows.
\IEEEraisesectionheading{
  \section{Introduction}\label{sec:introduction}
}
\IEEEPARstart{S}{olar}  irradiance, the output of light energy from the sun measured at a location on Earth, is converted to usable energy through the use of photovoltaic devices. The amount of energy received is usually affected by cloud or aerosol occlusion which scatters, reflects, or absorbs solar irradiance \cite{radiation, MissiontoPlanetEarthRoleofCloudsandRadiationinClimate, SERRANO201550, TZOUMANIKAS2016314}.

Predicting the amount of energy to be received is difficult due to the seemingly random shape and trajectory of clouds which are inherently influenced by a variety of complex factors.
 This, in-turn, makes it difficult to predict precisely when there will be a loss of power due to a reduction of solar irradiance being received \cite{camb, SOVACOOL2009288, Australia}.
The occurrence of these rapid fluctuations poses significant challenges for power grid operators. It leads to voltage and frequency fluctuations, limited time to adjust between energy sources, and ultimately energy disruptions \cite{diagne2013review, kumar2016grid, impram2020challenges, infield2020renewable}.
As a result, full integration of solar into the electricity grid poses challenges, in part, due to the difficulty in forecasting this intermittent natural phenomena. 
By imaging clouds and estimating their motion, we can formulate this into an imaging problem that can precisely forecast when a cloud will occlude the sun. 
By translating the overall problem into one of  a trajectory forecasting, we can gain foresight into when there will be a cloud occlusion of the sun and predict the overall dip in the solar energy received at the ground.

% Paragraph 2
Cloud motion around a particular localized region can be monitored using sky images captured periodically by a wide angle {field of view} (FoV) imager, usually a hemispherical mirror or a fisheye lens.
By using various prediction methods, cloud motion for future time instances can be predicted using these sky images. 
However, wide FoV imaging systems provide non-uniform spatial resolution of the sky with a higher detail and resolution at the zenith (or directly overhead) and significantly lower-resolution near the horizon.
As a consequence, our ability to detect a cloud at the horizon is negatively influenced due to the lowered resolution; further, the lower resolution also implies that the estimates of cloud velocity is also poor for those at the horizon.
This inability to detect clouds at the horizon and estimate their motion eventually limits the time horizon over which we can make precise predictions of solar occlusion by a cloud and the irradiance measured on the ground.
%
%Also, it is not assumed that the sun is always positioned at the zenith of the imager. The important aspect is the distance of the cloud to the line-of-sight from the camera to the sun. If the sun is closer to the horizon, then it could be that clouds that are directly above the imager moving towards the line-of-sight to the sun is what is important for accurate prediction. However, we make the case that traditional imagers already capture full resolution at the zenith and therefore, the focus should be further out clouds near the horizon.
%
%It is also likely that evaporation/condensation puts a cap on achievable prediction accuracy. However, we do not see a clear avenue to separate the effects of resolution at the horizon from non-uniform spatial resolution.
%
%As a result, these imagers are restricted in their ability to  long-term predictions of cloud motion due to their lack of angular resolution near the horizon. 
%
%This results in a non-linear mapping of motion in the image space which heavily compresses objects at the periphery of the imager. 	

\begin{figure}
\centering
\includegraphics[width=0.5\textwidth]{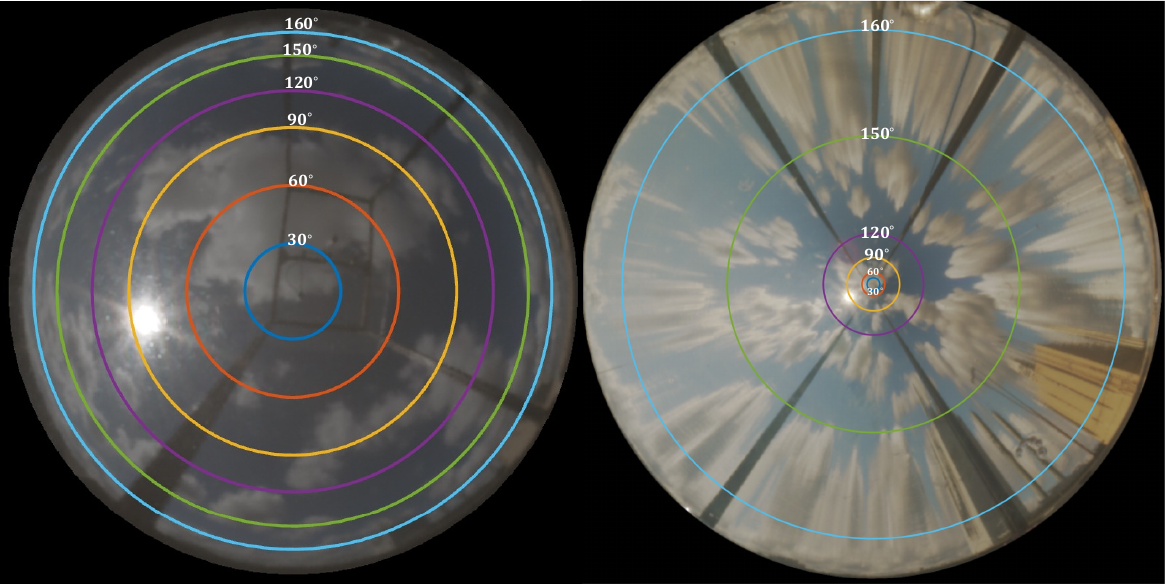}
\vspace{-22pt}
\caption{Our work presents a computational imaging system based on a catadioptric combination of mirrors and cameras. Seen above are images of the sky when using  (left) a traditional hemispherical mirror and (right) the proposed hyperboloidal mirror. Overlaid on top are circles corresponding to different angular extents of the sky.  The use of a hyperboloidal mirror enables increased. angular resolution near the horizon. We also propose a learning-based framework for predicting future solar irradiance. Together, these contributions enable a more accurate prediction over a longer time horizon than the  hemispherical imagers. }
%\vspace{-22pt}
\label{fig:grabber}
\end{figure}

As a result of these limitations, traditional works have attempted to combat this issue by digitally undistorting the sky image to limit the non-uniform spatial resolution \cite{Julian_2021_ICCV, su9040482, app11115049} or by employing a multiple camera setup \cite{Brown2007AutomaticPI, articleStitch, s23239481}.
Digital warping does help alleviate some of the challenges underlying non-uniform flow estimation, but it is fundamentally limited by the loss of resolution at image formation.
Adding more pixels by using more cameras or even a higher resolution sensor can be an effective approach, but comes with increased costs.
Further, direct imaging of the sky needs to be done with some care, given that the potential damage to the sensor caused by a focused image of the sun.
We instead pose a different question: is it possible to \textit{optically redistribute} the pixels in a wide FoV camera so that resolution is uniform for a cloud as it traverses the field of a sensor?

The centerpiece of this work is a catadioptric system that \textit{optically} warps the sky to provide a uniform spatial resolution of the sky (for each height), over the entire field of view of the camera.
We achieve this by imaging the sky through a  mirror whose shape is designed to provide the aforementioned property.
This design also has the added benefit of making motion of the clouds equally perceptible, be it at the zenith or the horizon.
As a result of using this mirror shape in a catadioptric setup, our ability to estimate cloud trajectory is improved over traditional methods even when a cloud is farther away. 
This improves long-term cloud evolution prediction and as a result, prediction of when a cloud will occlude the sun.
%
%To summarize, \textit{seeing farther in space enables predicting further in time!}
%, with a customized mirror shape that is  to provide uniform spatial resolution of the sky

{\flushleft \textbf{Contributions.}} Our method advances long-term forecasting of cloud evolution, enabling predictions that extend far beyond previous works, reaching into the tens of minutes. Our contributions are as follows:
	
	\begin{itemize}[leftmargin=*]
            \item \textit{Imaging system for whole sky imager}. We have designed and deployed a novel sky imager comprising a catadioptric system with an adapted hyperboloidal mirror \cite{710698} to capture and analyze sky images with the eventual goal of improved solar irradiance forecasting.
            \item \textit{Dataset of sky images}. Using our imaging system, we have captured high dynamic range sky images across a period spanning many months. The dataset also includes time-synchronized ground solar irradiance values captured via a pyranometer. 
             \item \textit{Predicting from spatial-temporal slices.} We have a novel prediction algorithm that uses estimated wind velocity to identify an informative 2D space-time slice of the imagery; this allows us to ignore the clouds that are unlikely to occlude the sun at our vantage point. More importantly, it significantly simplifies the resulting prediction problem, which we address using a lightweight learned-network.
        \end{itemize}

{\flushleft \textbf{Impact.}} The contributions above have resulted in a system that provides precise prediction of sun occlusion and solar irradiance over a time horizon of tens of minutes ($\sim$30 minutes for the simulated data, and 10-20 mins for the real system).
This \textit{long-term prediction} is an order of magnitude improvement over previous works; for example, Julian and Sankaranarayanan \cite{Julian_2021_ICCV} who rely on a similar premise, but with digital warping, show prediction results that only span 2-3 minutes.
Finally, we also hope this work will continue the renewed interest in this problem that is at the intersection of imaging and solar prediction.
To this effect, we have released the dataset and its associated code base   at \cite{SkyCamwebpage}. 
The underlying system is currently operational, and the size of the dataset will progressively increase over time. 
        
{\flushleft \textbf{Limitations.}} Our current prediction system is based on an assumption that clouds move linearly, driven by wind.
While this is true for the most part, at least over the half-hour or so of our prediction horizon, the model does not account for  cloud genesis and extinction events within the field of view.
It is very likely that such evaporation and condensation provides a fundamental limit on achievable prediction accuracy.
% However, there is no  clear avenue to separate the effects of resolution at the horizon from non-uniform spatial resolution.
%
Our real dataset includes such phenomena; this suggests the need for more robust statistical models along with additional data such as humidity and other atmospheric conditions to account for such events.
%
%The proposed setup for our catadioptric sky imager poses a number of limitations. Initially, the hyperboloidal mirror shape requires custom and precise fabrication processes. Consequently, giving the metal a highly reflective mirror surface has to either be done by manually sanding the surface or by a custom reflective coating. The use of hyperboloidal mirrors in a catadioptric system must also be accurately calibrated in terms of distance from the pinhole to prevent unwanted optical artifacts such as distortion and focal blur.

\section{Prior Work}
\label{sec:prior}
\renewcommand{\paragraph}[1]{{\flushleft \textbf{#1}}}

\paragraph{Large Field-of-View Sky Imagers.} Devices used to acquire images of the sky---often dubbed \textit{sky imagers}---encompass a fisheye lens or a catadioptric combination of lenses and hemispherical mirrors \cite{TSI, s23177343, 7455105}. These images are wide FoV RGB images captured in regular intervals. 
	
A limitation of these imagers is the non-linear fisheye distortion introduced which affects clouds optical flow estimates for trajectory predictions. The further-out clouds attenuate evolution over time and their prior motion estimates attribute to longer forecasting horizons. To combat this issue many works have attempted to spatially warp these images to achieve uniform apparent motion and limit the fisheye distortion \cite{PALETTA2022119924, su9040482, app11115049}. Julian and Sankaranarayanan \cite{Julian_2021_ICCV} have even shown that spatially warping these images achieves longer forecasting horizons. Limiting the long-term prediction accuracy when using digital warping is the loss of resolution of pixels at the periphery \cite{1225480, geom}. Think of it as digital zoom versus optical zoom. Periphery pixels are stretched out and then interpolated. Therefore, true pixel values at these locations are absent. For learning and optical flow based methods, these pixel values are essential for accurate long-term prediction.

%\subsection{Predicting Cloud Movement}
\paragraph{Predicting Cloud Movement.} Modeling cloud evolution through tracking and forecasting clouds solely using sky images are achieved using mainly two methods. Initial works utilized optical flow based methods \cite{8297111, 7455105, Jayadevan2012ForecastingSP, inproceedingsLowCost} using pairs of subsequent sky images to forecast cloud trajectory. Overall, this method is inadequate for long-term prediction due to the variability of cloud shapes and trajectory between image captures making it difficult to forecast. 
	
More recent works have seen greater success using deep learning to predict a subsequent sky image for a future time instance \cite{Julian_2021_ICCV, PALETTA2022119924, 7007633, articleCLoudPred, nie2023skygpt}. Learning-based methods can be further improved for more accurate long-term forecasting by addressing the fisheye distortion of the scene introduced by traditional large-FoV imagers. %This issue was previously addressed through digital warping \cite{Julian_2021_ICCV, 8549649}.
In this work, we show that optical warping leads to even greater success when coupled with learning-based predictions.

\paragraph{Photovoltaic Power Output Prediction.} 
Directly predicting photovoltaic power output has been a direction taken by previous works. These studies either take a statistical approach to predicting future irradiance values from past values \cite{SHARIFZADEH2019513}, \cite{ALZAHRANI2017304} or finding the relationship between an associated sky image and its irradiance value; also known as nowcasting \cite{9054985}, \cite{ZHANG2018267}. However, these works do not take the future state of cloud patterns into consideration which directly influences the amount of irradiance received at the ground. Therefore, by having an accurate method of predicting the future distribution of clouds, a better estimate of future irradiance can be obtained.

%In recent years, there has been a broad increase in the interest of utilizing computational imaging and learning-based techniques for solar irradiance forecasting *CITE*. These works either predict solar irradiance directly using statistical methods, captured sky images, or learning-based architectures which use a combination of various data sources.

\paragraph{Computational Imaging for Atmospheric Tomography.} 
The ideas in this paper are closely related to a rich body of work related to computational imaging of clouds, especially in the context of tomography reconstruction.
%
%Many works attempt to image clouds as a true three-dimensional (3D) volumetric matter rather than two-dimensional (2D) beings in images. These works fall under the category of solutions with the goal of tomographic reconstruction. 
This includes results that model the heterogeneous multi-scattering media and reconstruct the full volumetric field using distributed ground-based camera systems \cite{101007}, \cite{7492869}, \cite{MEJIA2018287}, \cite{9105241}, or airborne imagery \cite{1JRS12042603, 701027, amt-6-2007-2013, Ronen_2021_ICCV}. 
Such three-dimensional analysis of clouds also provides a pathway for prediction of solar irradiance over a large geographic region, provided we are able to recover the cloud distribution over the region; in contrast, we rely on simpler two-dimensional reasoning in our work given that we are making predictions at a single location that is collocated with our imaging instruments.

%\subsection{Catadioptric Imaging Systems}
\paragraph{Catadioptric Imaging Systems.} Catadioptric imaging which utilizes the reflective nature of mirrors during the acquisition process are designed such that their shape can achieve various tasks. Baker and Nayar \cite{710698} have extensively studied the family of these shapes. In particular, the hyperboloidal shape provides practical wide-angle imaging with minimal distortion and solves to the aforementioned challenges. This selected mirror shape will be further discussed in the subsequent sections.

%\acs{ASWIN add more relevant papers for catadioptric imaging}
%test

\section{Problem Definition}
\label{sec:problem}
In this section, we introduce the problem of sky imaging by stating the desired specifications and discussing the gaps between these requirements and current wide FoV imagers.

\subsection{Design Specifications}
Prediction of cloud movement requires precise estimates of their velocities when they are at the periphery of the FoV. 
For example, a  cloud with a typical height of 1 km and a velocity of 50 km/hr, will cover an angle of $\tan^{-1}(25/1) =87^\circ$, in a camera's view, over half an hour. 
If the sun is at the zenith, to provide a reliable prediction half-an-hour in the future,  we need a $174^\circ$ field of view, as well as the ability to precisely sense motion when the cloud is near the horizon.
This provides us with our design specifications: an imager with extremely large field of view approaching $180^\circ$, while providing the ability to detect and estimate cloud motion over the entire field.
We interpret the second part of the specification as one of providing uniform spatial resolution for the cloud as it appears and traverses the field of view of the imager.
While uniform resolution by itself is not a necessary condition (for example, we could ask for higher resolution at the horizon over the zenith), it allows for a robust solution that can also accommodate cloud creation events within the field of view.

\begin{figure}[tb]
            	\includegraphics[width=0.475\textwidth]{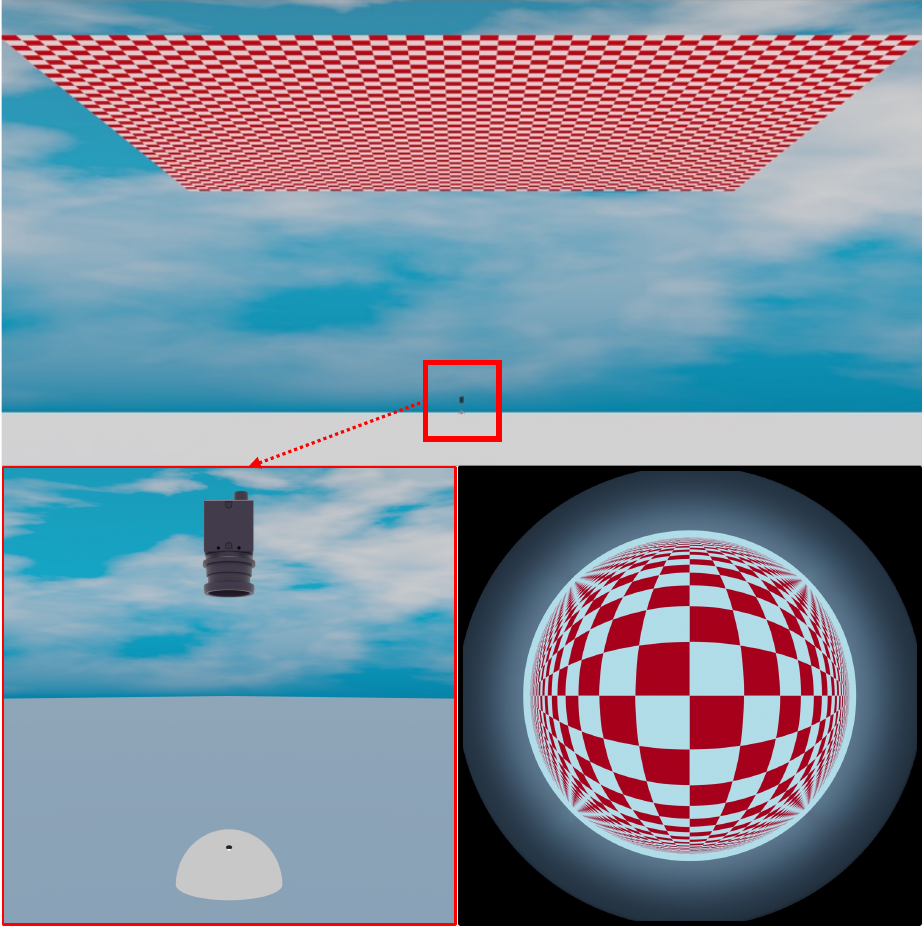}
	 \caption{A rendering obtained using Blender to visualize the non-uniform resolution of a hemispherical mirror. (Top) We create a scene consisting of a checkerboard with a length and width of 50 km, placed 2 km high above the ground. Each square on the checkerboard has physical extent of 1 km. (Bottom-left) Our imaging system consists of  a pinhole camera observing the sky or the checkerboard indirectly through a hemispherical mirror. (Bottom-right) The image observed on the camera has high resolution at the zenith of the image and significantly lower resolution at the periphery.}
	\label{fig:checkerboard_sphere}
\end{figure}

\subsection{Gaps in Current Sky Imagers}

Current wide FoV imagers can be built with a fisheye lens or more commonly with a catadioptric system where the sky is imaged through a hemispherical mirror.
Such traditional sky imagers are not conducive for long-term prediction due to their lack of  resolution at the periphery of the imager. 
%
%As we argued earlier, being able to predict the occlusion of the sun at future time instants requires us to detect clouds as and when they appear at the horizon.
%
%	To predict cloud motion further in time, we need to look fa back in space and this is drastically limited using a hemispherical mirror.
%

Specifically, in such systems, an object placed at the zenith of the sky will appear to have a larger total spatial extent as opposed to the same object at the horizon. 
Figure \ref{fig:checkerboard_sphere} visualizes this circumstance via a large checkerboard placed above a simulated hemispherical mirror. The checkerboard, which has a length and width of 50 km, is placed 2 km high above the mirror where each square is uniformly spaced at 1 km per square space. This hemispherical setup shows that squares at the zenith of the imager appear larger compared to squares at the periphery, despite their physical dimensions being the same. This compression of the squares at the horizon translates to poor localization of clouds that are located at the horizon in the world. 
Another related factor is our ability to estimate motion. In current sky imagers, motion of clouds appear to be non-uniform despite their physical speed being largely the same (since clouds are driven by wind), with large apparent motion at the zenith and significantly smaller ones at the horizon.
In more practical terms, the  imagery only allow for precise estimates of cloud velocity only after it is significantly away from the horizon; in turn, this limits the time horizon over which sun occlusions can be predicted.
%
%The compression of spatial features at the horizon has the added effect of making motion unobservable at the horizon.
%
%Uniformly imaging the motion of further-out clouds is necessary. If the model has an accurate understanding of how the cloud evolved in the past, it can more better attenuate its evolution and make confident long-term predictions into the future. Aforementioned, and clearly visualized in Figure \ref{fig:opticalFlow}, the non-uniform apparent motion of the clouds throughout the view of the imager is a direct result of the lack of pixel resolution at the boundary; ultimately limiting the long-term forecasting of cloud evolution.

\subsection{Solution Outline}

Our goal is to address the limitations of current sky imagers which  achieve a large viewing angle at the cost of two issues that limit long-term cloud motion prediction: lack of resolution at the horizon, and non-uniformity of motion.
How can the problem of non-linear motion and lack of pixel resolution be circumvented? 

Our approach relies on the insight that we can redesign the mirror used in a sky imager to spatially redistribute the pixels with the eventual goal of having the same spatial resolution on a cloud over the field of view of the imaging system---immaterial of whether the cloud is at the horizon or at the zenith.
This allows for early detection of clouds, as well as simplifies the motion estimation problem since the clouds largely translate over the field of view.
We discuss the mirror design problem in Section \ref{sec:cata} as the associated testbed and dataset in Section \ref{sec:skycam}.

The second part of our solution is an algorithmic technique for prediction over time-horizons of tens of minutes.
Part of the challenge here is the high-dimensionality of the input image which makes any learning-based solution hard to implement due to compute and memory requirements, as well as the need for a large amount of input data.
To simplify this problem, we argue that cloud motion due to wind is largely translational; hence, to predict the occlusion of the sun as well as solar irradiance at future time instants, it is sufficient if we look at a spatial slice through the sun that is parallel to the wind direction.
While this likely misses out on predicting irradiance due to indirect skylight, it has all the relevant information for predicting direct sunlight which is the dominant term in the overall irradiance. 
We describe this algorithm in Section \ref{sec:algo}.

Finally,  %we build and deploy a testbed, and have collected a dataset of sky images over a period spanning five months.
we evaluate our algorithms over this dataset, as well as a synthetic counterpart, in Section \ref{sec:eval}.

%Naively, we add more pixels. However, this increases computational costs and we still face the issue of non-linearity. Another solution is to digitally warp the image to a new space that achieves uniform apparent motion, but again, we face the issue of loss of pixel resolution during the interpolation step. We desire to address these constraints by designing a mirror that maintains the benefits of traditional sky imagers all-while limiting the traditional constraints. This mirror should maintain the large FoV imagery, uniform resolution, along with uniform apparent motion throughout the whole image. We now provide a brief analysis of this desired mirror shape and catadioptric setup.

\section{Mirror Design}
\label{sec:cata}

\begin{figure}[tb]
  \centering
    \includegraphics[width=0.25\textwidth]{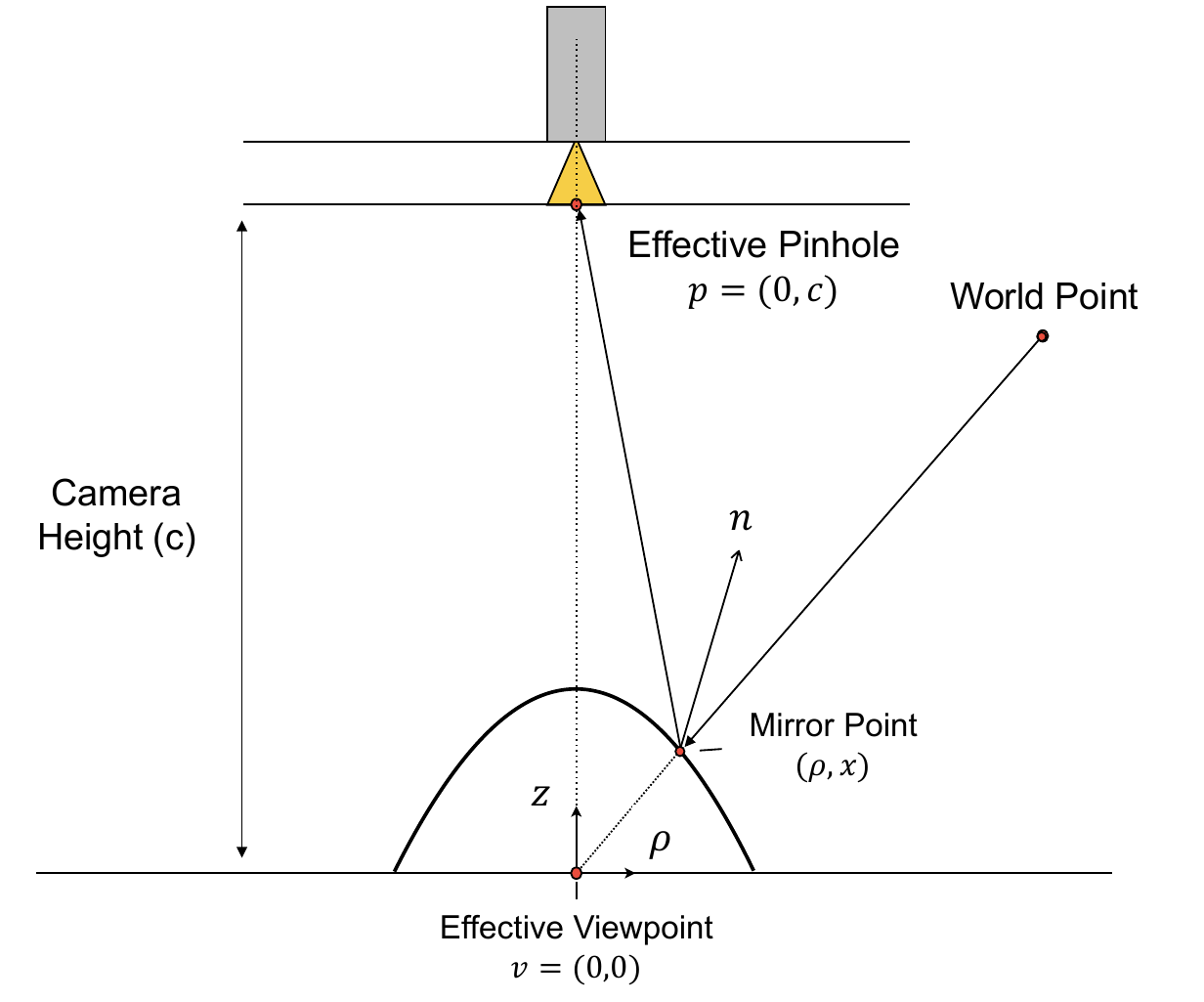}
	\includegraphics[trim={57em 5em 8em 6em}, clip, width=0.2\textwidth]{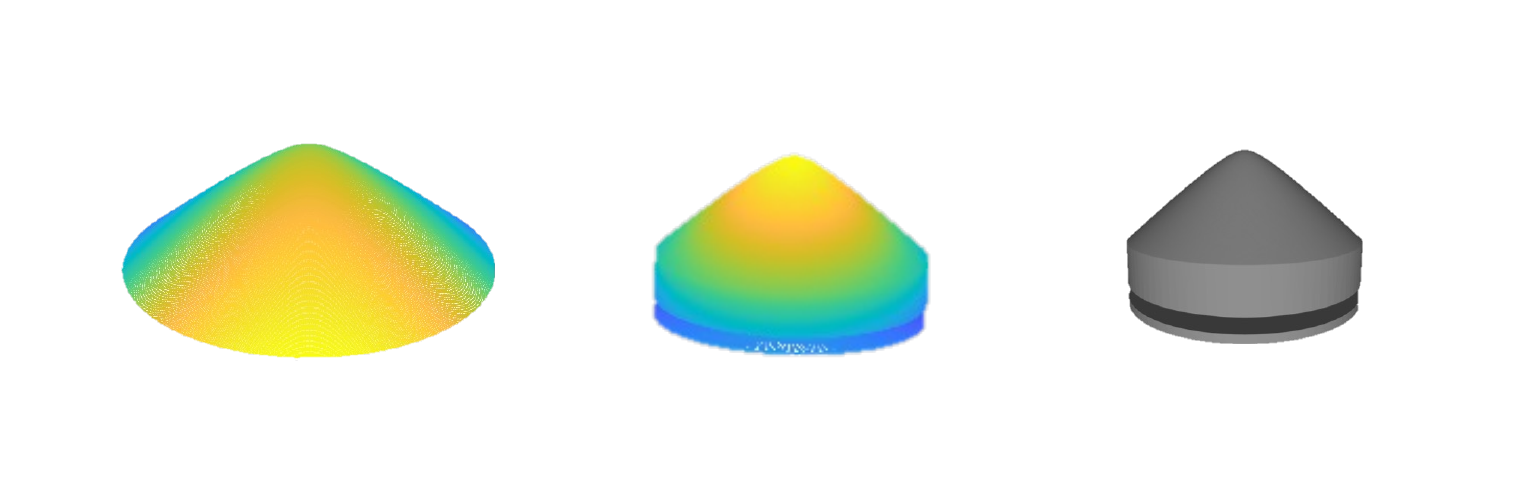}
  \caption{(Left) We design the mirror for a setup where a camera is placed 1 meter above the mirror. The shape is optimized so that the overall system scales the field of view of the pinhole camera from $3.58^\circ$ to $170^\circ$ uniformly. (Right) The resulting hyperboloidal mirror shape that we use in our setup.}
    \label{fig:equationAngles}
\end{figure}

%\subsection{Shape of the Mirror}
We frame the problem of mirror design as one that ``flattens'' the sky image formed on the sensor.
Figure \ref{fig:equationAngles} illustrates the relevant variables.

We delve into the derivation of the mirror shape profile, which share the same goals as \cite{710698}, in the supplemental material. However, we concisely present the setup here. Our basic setup is that of a pinhole camera with a sensor size of $w = 12.5$mm, placed at a distance $c=1$m from the mirror, with a field of view of $f_c=200$mm. These choices are based on design considerations for the final implementation where we need the camera to be sufficiently far away to avoid blocking a significant portion of the field of view. The long focal length also allows us to effectively mimic the pinhole camera with a lens-based counterpart.

The mirror has a shape $z = f(\rho)$, where $\rho$ is the radial distance over the ground plane.
We make an additional assumption that the cloud is at some height $h$; the exact height of the clouds do not play an actual role as we will assume that $h \gg c$ and so only the tangent of the angle subtended by the cloud at the mirror matters.
With this, we formulate the mirror design as one of designing the profile $f(\cdot)$ such that the effective sky to sensor mapping is a  \textit{scaling operation over the desired field of view}.
Effectively, we are scaling the FoV of the camera---which is $\theta_{cam}=3.58^{\circ}$---using the mirror by a constant spatial factor to achieve a target FoV $\theta_{target}=170^{\circ}$.

To determine the mirror shape, we use a numerical procedure where we solve for the axial profile $f(\cdot)$ by densely ray tracing over the image plane. With this, we also have the constraint that the ray --- after mirror reflection --- behaves like a pinhole camera with the target field of view.
This provides a constraint on the derivative of the $f$ (since the surface normal is determined by the normal).
Integrating this derivative provides us with the desired shape.
A visualization of  the resulting mirror shape is shown in Figure \ref{fig:equationAngles} (right). 
This solution falls under the family of shapes described by Baker and Nayar \cite{710698}, and in particular it is a hyperboloidal shape.  

\begin{figure}[tb]
  \centering
    \includegraphics[width=1\linewidth]{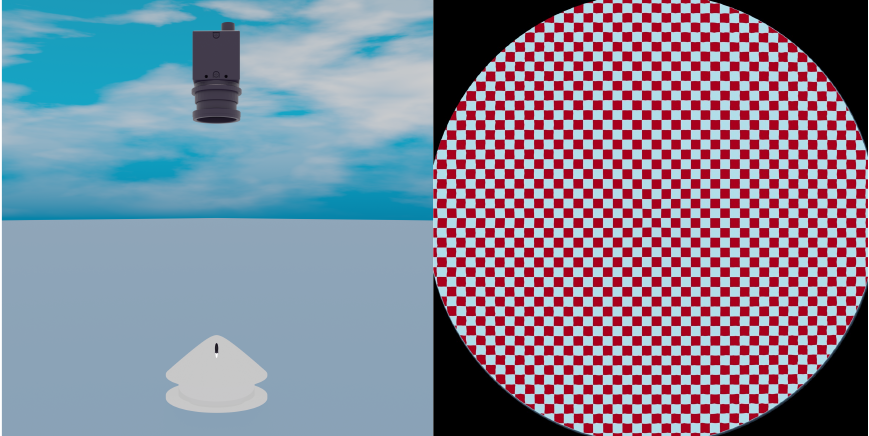}
  \caption{Simulated case visualizing the uniform resolution of our hyperboloidal mirror. (Left) The same parameters as Figure \ref{fig:checkerboard_sphere} with the proposed mirror replacing the hemispherical mirror. (Right) Observe how the checkerboard resolution is spatially uniform---a consequence of the system acting as overall scaling operation.}
    \label{fig:checkerboard_hyper}
\end{figure}

In Figure \ref{fig:checkerboard_hyper}, we use Blender to render an identical setup as Figure \ref{fig:checkerboard_sphere}, but with the hyperboloidal shaped mirror. Our mirror achieves a uniform image of the checkerboard while maintaining a large FoV, showing that we are able to image the sky with uniform resolution. As is to be expected, this design also enables uniform motion estimates throughout the whole FoV of the imager.
%\subsection{Analysis}
%
%\subsection{Slice-based Prediction}
%
%\subsection{Simulation Results}

\section{Testbed}
\label{sec:skycam}
% Synthetic Images Gallery

%% Smear Combined FIgures
%	\begin{figure}[t]
%                	\begin{center}
%%                   		\includegraphics[width=1\linewidth]{Figures/Smear_Combined_Figures}
%				\includegraphics[width=0.5\linewidth]{Figures/Smear_Cube_Hyper}
%
%                	\end{center}
%	                   	\caption{Visualizing the smearing caused by the hyperboloidal mirror. (a) Visualizes a cube at the zenith of the mirror which is not affected by the object height. (b) A thin cube placed 10km away from the mirror. (c) A Thick cube placed 10km away from the mirror.}
%		\label{fig:smearing}
%\end{figure}

We now describe our imaging setup in the context of our hyperboloidal-based mirror.

%% Wean Roof Setup
%	\begin{figure}[t]
%				
%	\centering
%		\begin{subfigure}[b]{\textwidth}
%%		\hspace{0.8cm} 
%		\includegraphics[width=0.5\linewidth]{Figures/SkyCam_Setup}
%		\end{subfigure}
%	
%	\caption{Image of our real-world acquisition setup. (Left) The complete setup. (Right-top) Detailed visualization of mirror placement of the hyperboloidal mirror and spherical mirror. (Right-bottom) Detailed view of the polished hyperboloidal mirror.}
%	\label{fig:weanSetup}
%\end{figure}

\subsection{Simulation Setup}
We initially evaluate and report results of our setup and methods on simulated data which achieves the idealized scenario of a real-world setup with known parameters. The platform used to develop our simulated data is Blender which uses the same setup as in our real-world data. In Blender, the Pure-Sky Pro package which simulates an array of cloud formations inspired by \cite{link} is used. Although not modeled as mathematically in-depth as a large-eddy simulation \cite{LES}, Pure-Sky Pro is accurate to the scale of this simplified simulated application. The package does allow for the modification of cloud dynamics such as how warm/cold air affects cloud evolution.

\begin{figure}[t]
	\centering
        			\includegraphics[width=0.155\textwidth]{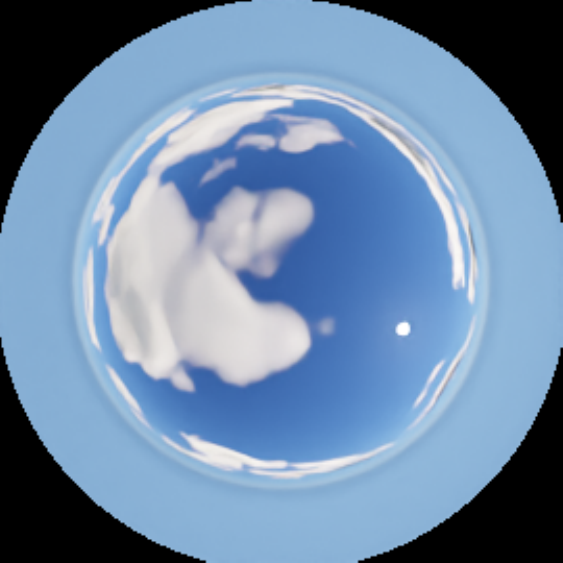}
        			\includegraphics[width=0.155\textwidth]{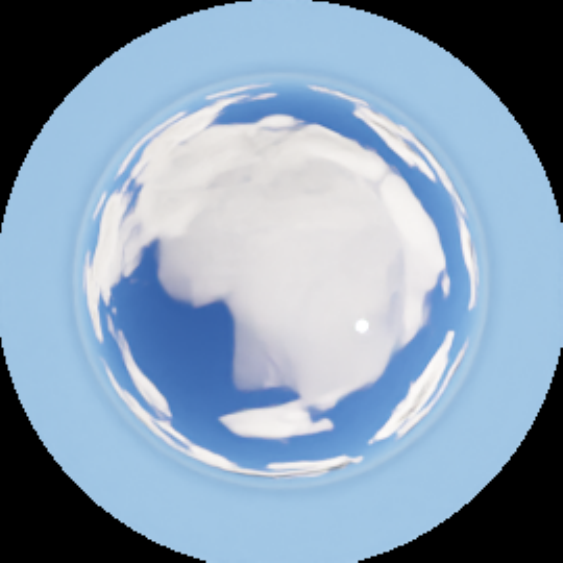}
				\includegraphics[width=0.155\textwidth]{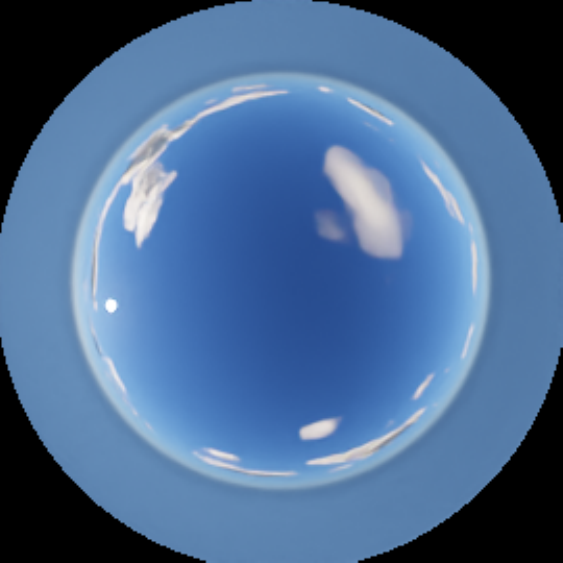}
        			\includegraphics[width=0.155\textwidth]{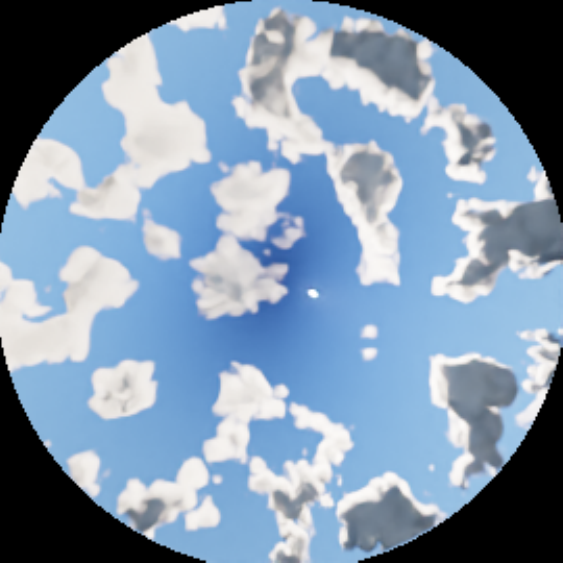}
        			\includegraphics[width=0.155\textwidth]{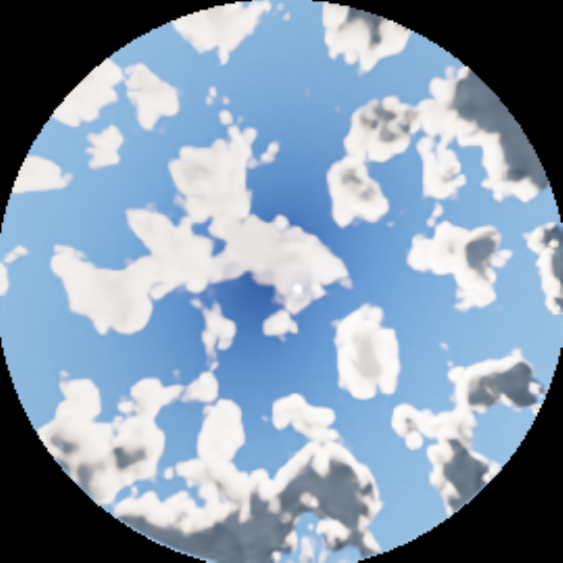}
				\includegraphics[width=0.155\textwidth]{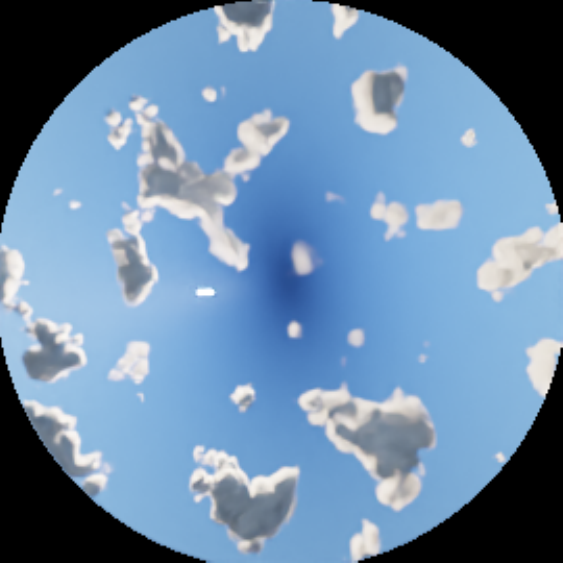}
	 \caption{Various images captures from the synthetic dataset. (Top) Captures from hemispherical setup. (Bottom) Captures from the hyperboloidal setup. Each column is captured at the same time instant}
	\label{fig:synthGallery}
\end{figure}

Using the computer generated hyperboloidal-mirror shape, as shown in Figures \ref{fig:checkerboard_sphere} and \ref{fig:checkerboard_hyper} placed with a reflective mirror material property, we capture simulated data on various cloud scenes with a sampling period of $T_0 = 30$ seconds. We also capture the same data using a hemispherical mirror with the same parameters. These cloud scenes include randomized cloud parameters across a 28-day period from 8AM to 5PM based on real-world factors such as wind, hot/cold air patterns, and sunlight. Images of simulated data for both mirrors are shown in Figure \ref{fig:synthGallery}. %Further details of our simulated setup are discussed in the supplemental material.

\subsection{Hardware Prototype}

To develop a physical prototype for our mirror, we fabricated the desired 3D surface of the hyperboloidal mirror using a Computer Numerical Control (CNC)  machine. 
%we computationally plotted the equation of a line for the mirror shape as described in Section \ref{sec:cata}. Due to rotational symmetry, a surface of revolution can be formed about the vertical $z{\text -}axis$. Now, armed with the mirror shape as a 3D surface, we can export the 3D shape to a format suitable G-code that can be used to fabricate metal using a Computer Numerical Control (CNC)  machine. 
%
We used aluminum for the material due to the ease with which it could be polished; for the final mirror surface we used  a chemical deposition process that is commonly used to produce highly reflective surfaces \cite{silver}.
%Figure \ref{fig:mirrorGeneration} shows the resulting hyperboloidal mirror with a reflective mirror surface.

\begin{figure}
\centering
		\includegraphics[width=0.475\textwidth]{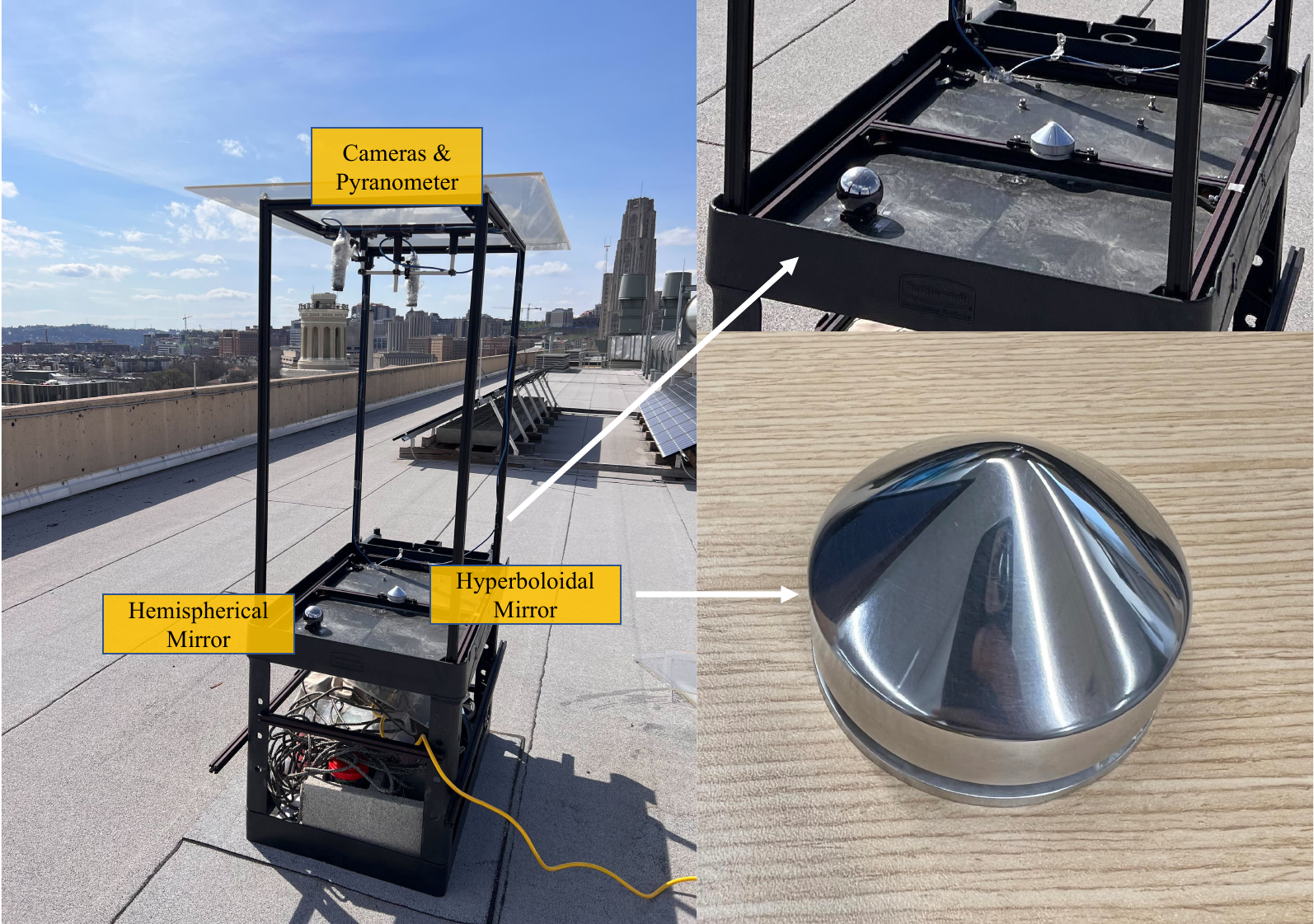}
\caption{Our deployed testbed with a  detailed visualization of mirror placement of the hyperboloidal mirror and hemispherical mirror.}
\label{fig:weanSetup}
%  \label{fig:opticalFlow}
\end{figure}

	For image acquisition in our system, we utilize an RGB camera %(FLIRBlackflySBFS-U3-200S6C)% 
	mounted on a cuboidal frame above the mirror.
	The mirror itself lies on the horizontal axis of the frame and coupled with a mini PC, captures sky images with a sampling period $T_0= 30 \text{ seconds}$. To minimize any nearby building occlusion, our imaging device is placed on a building roof and captures data continuously during daylight with a frequency of $T_0$. We also included a second system with a hemispherical mirror for evaluating improvements of our proposed work. 
To handle the large dynamic range of the sky, due to the sun, we capture  images using exposure bracketing and fuse them to get a single HDR image.
	The top of our system also includes a pyranometer that measures solar irradiance in the form of global horizontal irradiance (GHI), defined as the total  solar irradiance received at a location horizontal to the Earth's surface measured in the units of watts per meters-squared $\left( W/m^{2} \right)$. Our setup as described is shown in Figure \ref{fig:weanSetup}.

\begin{figure}[t!]
\centering
		\includegraphics[width=0.45\textwidth]{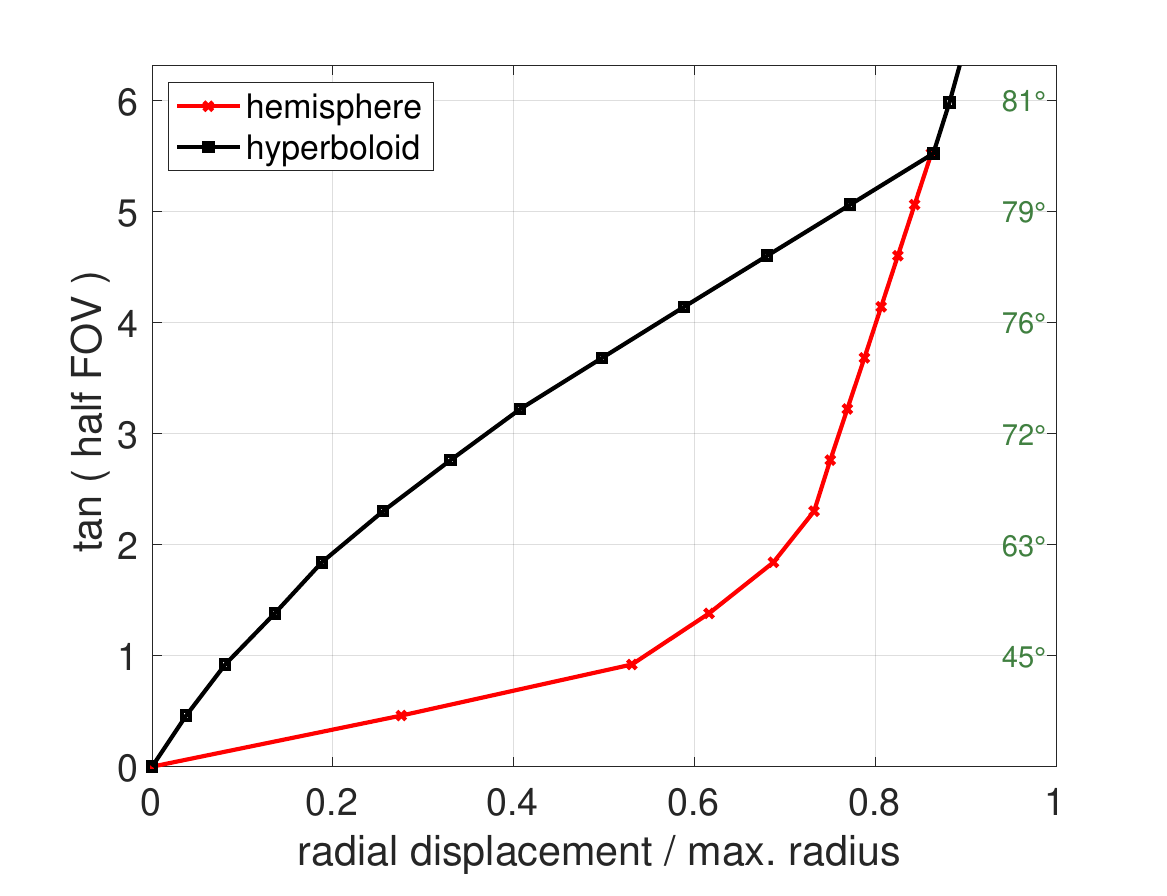}
\caption{Field of view of the sky (in tangent of angle) as a function of radial distance from center for the hyperboloidal and hemispherical mirrors in our deployed system. The numbers to the right of the figure provide the corresponding half-FoV in degrees. Note how the hyperboloidal mirror provides a linear relationship to the radial pixel displacement, thereby leading to flattening of the sky.
}
\label{fig:FOV}
%  \label{fig:opticalFlow}
\end{figure}

\begin{figure*}
	\centering
        	\includegraphics[width=1\linewidth]{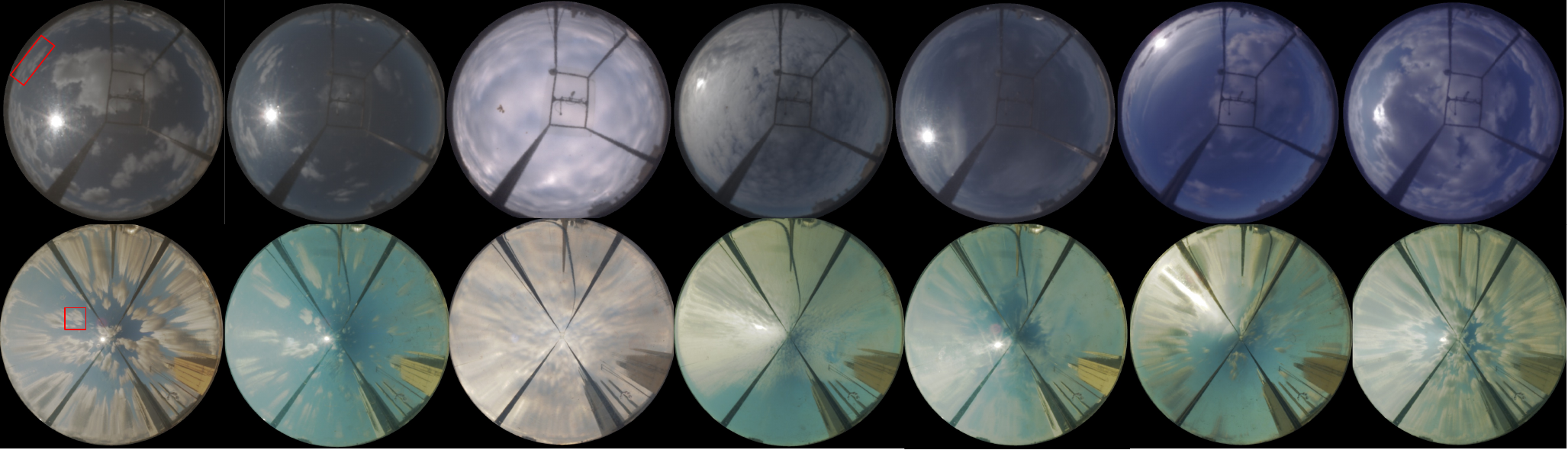}
	\caption{Real images captured from various dates and weather conditions. (Top) Images from the hemispherical setup. (Bottom) Images from the proposed hyperboloidal setup. Each column is captured at the same time instant. The cropped cloud in the red box in both images show the benefit of our mirror being able to image a cloud much further out.}
	\label{fig:realImgGallery}
\end{figure*}

\begin{figure}
\centering
            	\includegraphics[ width=0.475\textwidth]{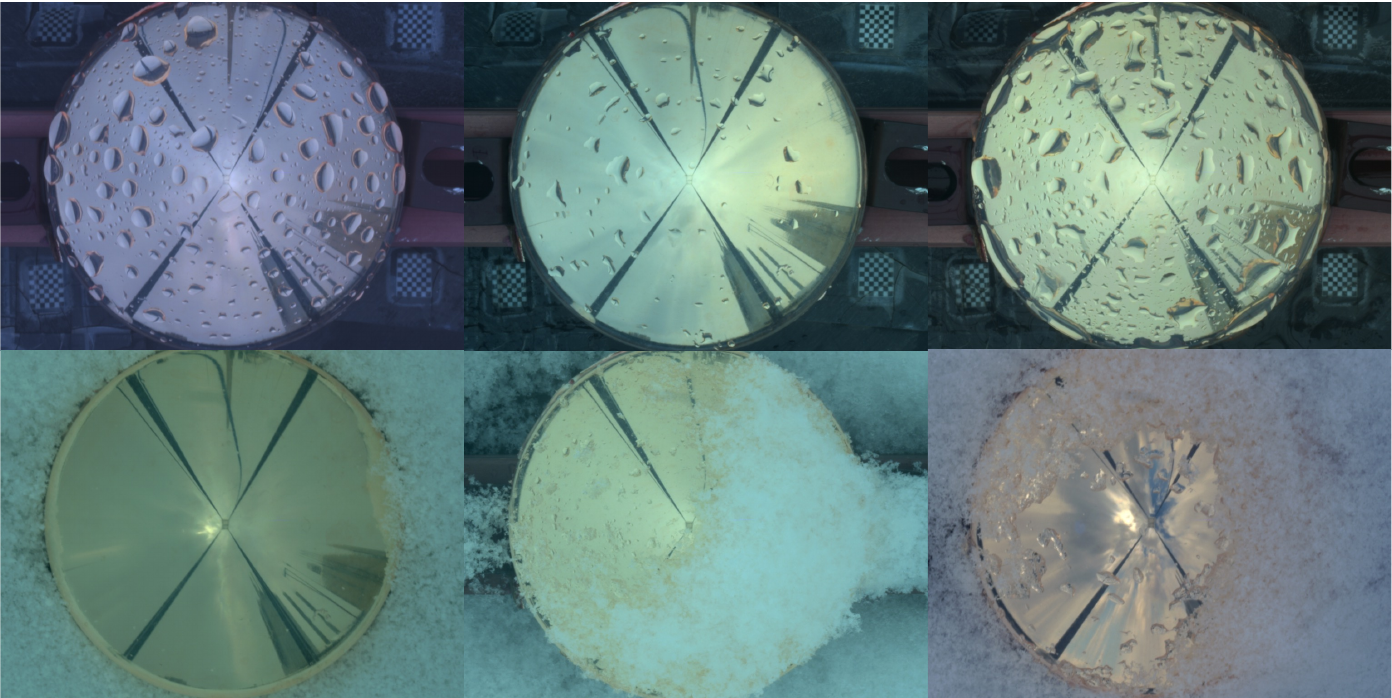} 
	\vspace{5mm} \\
	 \caption{We show examples of adverse weather conditions captured from our real setup of rain (top) and snow (bottom).} 
	\label{fig:other_conditions}
\end{figure}

Next, we characterize the angular mapping provided by both mirrors in our system in Figure \ref{fig:FOV}.
%
%For the real setup, we measured the FoV by placing a light source at various distances from both mirrors. We show the comparison between the hemispherical and hyperboloidal mirror in Figure \ref{fig:FOV}.
%
We placed a light source at various heights along the frame of our real setup, and used it to calculate the FoV observed at each mirror as a function of radial distance from the center of the mirrors, as observed in the acquired images. 
Figure \ref{fig:FOV} plots the half FoV, in its tangent, observed at each mirror across the image; for consistency, we normalize the radial distance of each mirror  by the radius of the mirror for a fair comparison.
Two key observations emerge. 
First, the hyperboloidal mirror provides a linear mapping between the radial distance observed in the image plane and the tangent of the observed half FoV; this is a key design property of the mirror that allows a fixed plane above the ground to be mapped to the image plane of the camera.
Second, notice how the hemispherical mirror devotes most of the image plane to a small central cone at the zenith of the sky.
Specifically, the central  cone of $45^\circ$ around the zenith of the sky occupies more than 50\% of the radial distance  with the hemisphere as opposed to less than 10\% with the hyperboloid.

\subsection{Dataset Collection}
%% Real Dataset Gallery
%\begin{figure*}
%	\centering
%        	\includegraphics[width=1\linewidth]{Figures/Real_Images_Dataset/Real_Dataset_Gallery2}
%	\caption{Real images captured from various dates and weather conditions. (Top) Images from the hemispherical setup. (Bottom) Images from the proposed hyperboloidal setup. Each column is captured at the same time instant. The cropped cloud in the red box in both images show the benefit of our mirror being able to image a cloud much further out.}
%	\label{fig:realImgGallery}
%\end{figure*}
	%1. Cite figures of images that were collected and show a gallery. Show images of sphere, optimal, and sphere warped images \\
	Figure \ref{fig:realImgGallery} shows a gallery of real images captured from our setup. These images are captured from our hyperboloidal mirror and a hemispherical mirror at the same time instant. Similar to the simulated data, the real images achieve the benefits of using the hyperboloidal shaped mirror. We are able to see more clouds within a single capture and the motion is more translational through time.
	Our dataset for this work consists of imagery collected from October 20th, 2023 to March 5th, 2024. We excluded days that were entirely cloudy or completely clear skied, so as to remove scenarios where GHI is nearly constant over the entire day.
This left us with 76 days worth of data with most days having partly cloudy conditions.
In Figure \ref{fig:other_conditions}, we show some adverse conditions from the captured datasets.

%\begin{figure*}
%	\centering
%        	\includegraphics[width=1\linewidth]{Figures/Real_Images_Dataset/Real_Dataset_Gallery2}
%	\caption{Real images captured from various dates and weather conditions. (Top) Images from the hemispherical setup. (Bottom) Images from the proposed hyperboloidal setup. Each column is captured at the same time instant. The cropped cloud in the red box in both images show the benefit of our mirror being able to image a cloud much further out.}
%	\label{fig:realImgGallery}
%\end{figure*}

	% *Add a subnote referring the reader to the supp material for a gif of the data*, and there is no need to warp the images to a different space. 

%
%\subsection{Simulation testbed}
%
%\subsection{Real Testbed}
%
%\paragraph{Mirror fabrication}
%
%
%\subsection{Dataset}

\subsection{Pre-processing}
\label{sec:preprocess}
Before we can apply learning-based techniques on this dataset, we need to perform certain operations on it. In particular, knowledge of the sun as well as the  wind velocity at each frame is helpful for the algorithms we describe next.

\paragraph{Sun localization.}
As a crude pre-processing step, we use the shortest exposure in our HDR stack to estimate the location of the sun.
However, this technique fails when the sun is occluded by clouds.
To get a robust estimate, we pool the data across multiple days of maximal saturation and reject outliers using RANSAC. This provides a sun estimate as a function of daytime where occluded sun estimates are filled by fitting a polynomial function over sun locations identified by maximal saturation.
Of course this will fail for incorrect predictions; therefore manual identification is required for some cases.

On an aside, the location of the sun in absolute angular coordinates with respect to the zenith of the sky can be analytically computed given the latitude and longitude of the testbed.
We can in principle map such elevation and azimuthal position of the sun to the image plane coordinates using a  calibration procedure; details of such a procedure can be found in prior work on cloud imaging \cite{9105241}.
%
%We opted for a simpler approach which does not require what we felt was a complicated calibration problem that had to account for the mirror shape.

\paragraph{Wind velocity estimation.}
Another useful information for the learning-based formulation that we will present next is the direction of wind velocity.
A challenge here is that clouds are largely featureless, which makes traditional optical flow techniques fragile.
Further, there are features in our field of view which are constant, for example buildings at the periphery and the frame used to hold the cameras.
These static features bias the optical flow estimates especially since they are also high contrast ones.

To overcome these effects we use the mask to suppress the static regions and run the optical flow technique proposed by Liu \cite{liu2009beyond} with very a strong weight associated with the spatial regularization term.
Finally, we use an aggressive temporal median filter on the estimated optical flow across frames to ensure a smooth flow field.

\section{Algorithms for Long Time-Horizon Prediction}
\label{sec:algo}

%    \begin{figure}
%    \includegraphics[width=1\linewidth]{Figures/NL_occ_steps.pdf}
%    \includegraphics[width=1\linewidth]{Figures/keogram_steps.pdf}
%   \caption{(Top) Space-time image. (Middle-left-to-right) We visualize the non-learning steps for occlusion prediction. We take some window out of the original space-time image, warp it based on the optimal $\hat{\theta}$, convert it to a new color space based on \cite{HYTA} and plot the mean through time. Red points show ground truth occlusion states. (Bottom-left-to-right) We show how space-time-slices are extracted based on the wind direction. The bottom far right shows the space-time image produced by the hyperboloidal mirror on the top and the hemispherical mirror on the bottom.}
%    \label{fig:keogram_steps}
%%  \label{fig:opticalFlow}
%\end{figure}

%%

The sudden rise and fall of received solar irradiance at the ground within a short period of time is crucial information for electricity grid operators to mitigate disruptions in power output. This event, also called a ramp event (RE) \cite{CUI2017227, godfrey2010modeling}, is influenced by the occlusion state of the sun by a cloud. This is necessary to forecast and can be achieved through prediction of cloud trajectory. Thus, we show the benefits of utilizing our hyperboloidal imaging system and focus on a suite of  algorithms that can exploit these benefits to better predict RE's.

\begin{figure}[tb]
  \centering
      \includegraphics[trim={0 0 49.5em 0}, clip, width=1\linewidth]{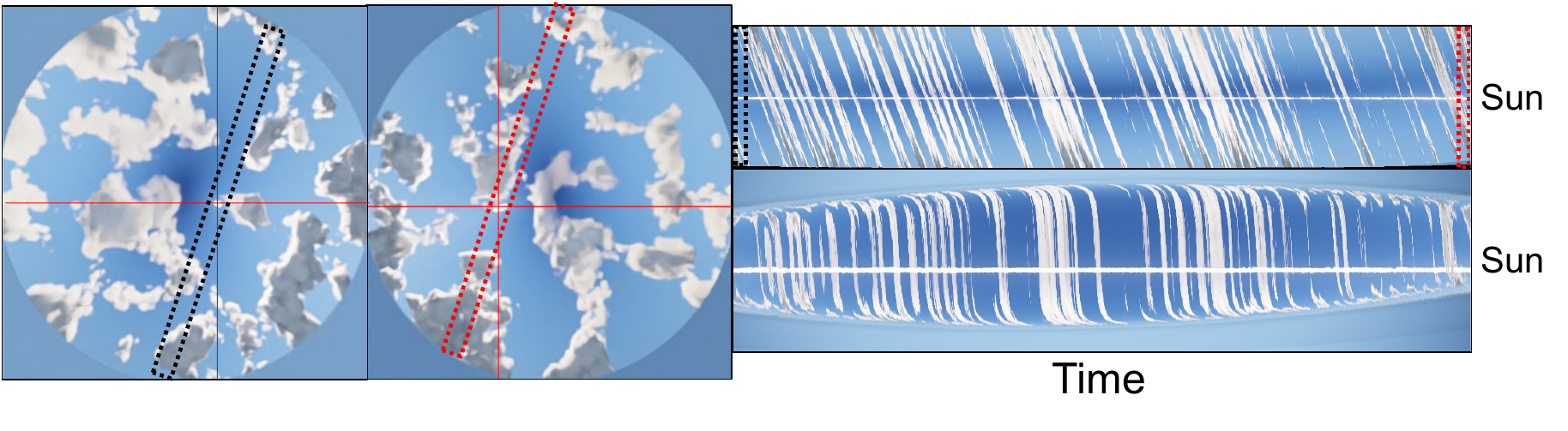}
      \includegraphics[trim={43.2em 0  1em 0}, clip, width=1\linewidth]{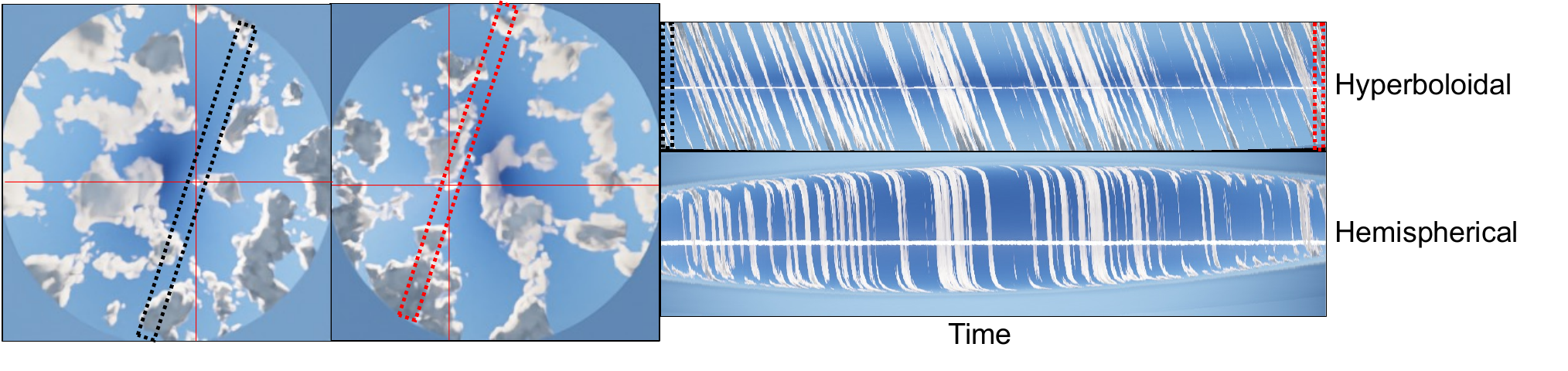}
    \caption{ We show how space-time-slices are extracted (top) Images of the sky from the hyperboloidal mirror at the start and end times of a sequence. We track the sun location and slice the image along the direction of wind motion, as marked in each image.
    (bottom) The space-time image produced by the hyperboloidal mirror on the top and the hemispherical mirror on the bottom. }
    %    We show that a cloud at time instance $\tau$ will occlude the sun at time instance T based on some angle $\theta$. If we take use this fact, we can take this cloud at $\tau$ and warp it to where it will be in the future at T+N. This is the intuition behind our non-learning approach and how we warp the images as input to our learning-based models.}
    \label{fig:idea}
    \end{figure}

%\subsection{SkyNet}
% We initially perform future frame prediction on the sky images. Images captured from both the hemispherical imager and our current hyperboloidal setup are used for this step. 
%	
%	Using the SkyNet \cite{Julian_2021_ICCV} model as a future frame predictor, we concatenate 4 images $\{ I_{t-5}, I_{t-3}, I_{t-1}, I_{t}\}$ , representing past time instances, as input into the model to predict an image at a future instance $\hat{I}_{t+1}$. To predict longer into the future, we recursively concatenate 4 frames using the associated time instances as input into the model $\{ I_{t+n-6}, I_{t+n-4}, I_{t+n-2}, I_{t+n-1}\}$ to predict an image $\hat{I}_{t+n}$. Further details of the SkyNet model can be referenced from the paper.
%	
%	For base inference, we predict to an image at $t+1$ and similar to the SkyNet paper, we use peak signal-to-noise ratio (PSNR) and the structural similarity index measure (SSIM) as metrics comparing future frames to its associated ground truth. Table \ref{tab:SkyNet_Results} shows these results for our hyperboloidal mirror and hemispherical mirror. Figure *CITE* shows side-by-side comparisons on predicted images for both setups.

\subsection{Space-Time-Slice Image}

\label{subsec:keogram}
The key benefit of having uniform apparent motion of clouds with the hyperboloidal mirror is that we can linearly back-trace the trajectory of clouds through time to  the cloud's projected path toward the sun.
 We can interestingly use this fact and state that the only part of the image that is important is the sun and clouds that are moving towards it. This corresponds to a linear slice of the image along the wind velocity as shown in Figure \ref{fig:idea}.
 By collecting such slices over time, centered around the sun as it moves, we can build a \textit{space-time-slice image} that effectively summarizes the relevant cloud movement.
 % We can disregard other parts of the image and simplify this problem. This will be done through the use of \textit{space-time-slices} which summarizes cloud patterns throughout a day by simply using a single image (see Figure \ref{fig:idea}). 
%A space-time-slice image takes a narrowband slice of a full sky image and horizontally stacks the slices through time to form a single image. Each slice is taken from a distinct image at time instance $t$ where the horizontal x-axis of the final space-time image represents time and the vertical y-axis represents space. 

We briefly describe how we create the space-time image and utilize it for inference.
For each time instant for a single day, the x and y coordinate of the sun is initially identified. Next, the general direction of cloud motion $(\hat{\theta})$ through time is obtained. We take a sun-centered slice of the image in the direction of cloud motion for each time instant and horizontally concatenate these images through time to formulate the space-time image. 
%Figure \ref{fig:keogram_steps} visualizes each described state and shows the resulting space-time-slice images. 
%
\par For the case of the simulated sky images, we are benefited by having the ground truth sun location, direction of cloud motion, along with the binary sun occlusion state. Therefore, we have the necessary parameters for an ideal space-time image. Figure \ref{fig:idea} shows this space-time image and compares the image obtained from our system to the hemispherical-based system. Our system clearly achieves the desired linear apparent motion and is the ideal case for predicting sun occlusion discussed in Section \ref{subseq:Non-Learning_Occlusion_Prediction}.
\par
\begin{figure}[tb]
\centering
            	\includegraphics[ width=0.475\textwidth, height=0.2\textwidth]{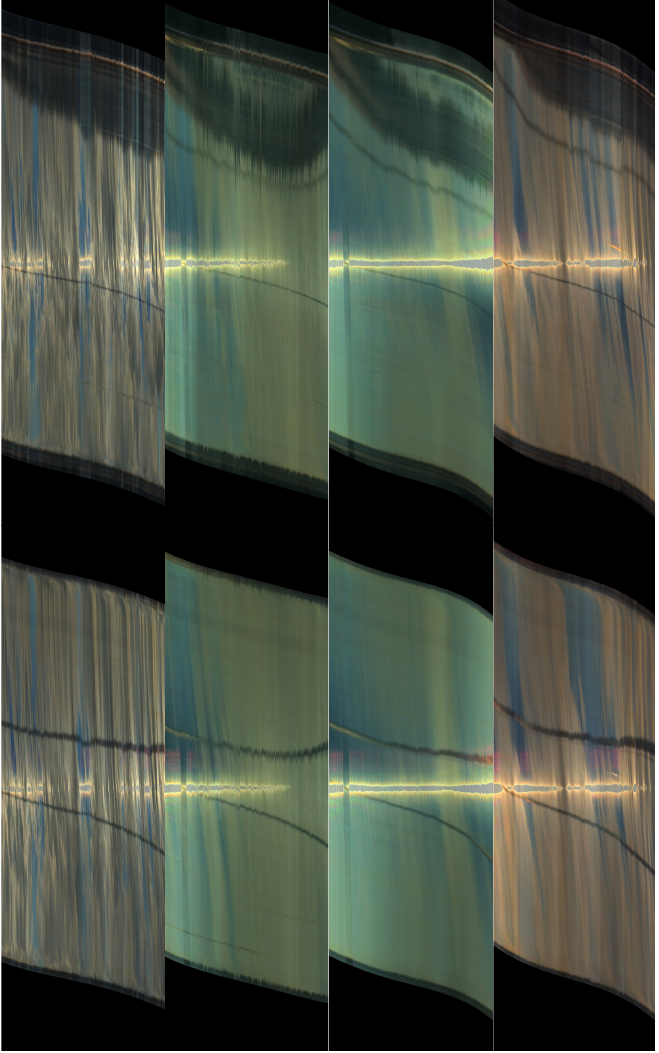} 
	\vspace{5mm} \\
	 \caption{We show resulting space-time-slice images from captured from our real setup. (Top) a result of selecting $\theta$ = $65^{\circ}$ while the bottom is $\theta$ = $85^{\circ}$.} 
	\label{fig:real_keo}
\end{figure}

 However, for the real images, ground-truth parameters are not given and are estimated. Sun location and general cloud direction is identified using the methods described in Section \ref{sec:preprocess}. 
In the real-world environment, cloud motion direction is not static and changes through time. Therefore, we estimate cloud motion over a whole day using estimated optical flow and apply an aggressive temporal median filter to ensure a smooth flow field.
We present sample space-time slice images for the real case in Figure \ref{fig:real_keo}.

%\subsection{Real Setup}

\subsection{Non-Learning Occlusion Prediction}
\label{subseq:Non-Learning_Occlusion_Prediction}
%%%Non learniing occlusion%%%%
%%%%%%%%%%%%

Given that the ground-truth parameters for the simulated images are available, we are able to perform non-learning based sun occlusion and show the benefits of our setup.

\subsubsection{Back Projected Sun Occlusion Prediction} 
\label{backproject}
Looking at the space-time image produced by our hyperboloidal shaped mirror, the linear streaks of clouds whose trajectory through time occludes the sun at the center of the space-time image, can easily be seen. Due to the non-linearity of the apparent motion within the hemispherical images, these space-time images do not have the same effect. We can use this and make the assumption that a cloud that occluded the sun at a time instant $T$ is the same cloud that is at a location $v$ on the image such that:
\begin{equation}
	v = \tau \tan \theta,
	\label{eq:keoEq}
\end{equation}
where $\tau = \left( T - t \right) $ is the time displacement from $T$ along the horizontal x-axis. If we sweep over a range of $\tau$ and obtain the slice warped to the current location of the sun at $T$ based on Eq.\ \eqref{eq:keoEq}, this results in a new image which, although seemingly meaningless, actually obtains information about the future sun occlusion states.

\begin{figure}
    \includegraphics[width=1\linewidth]{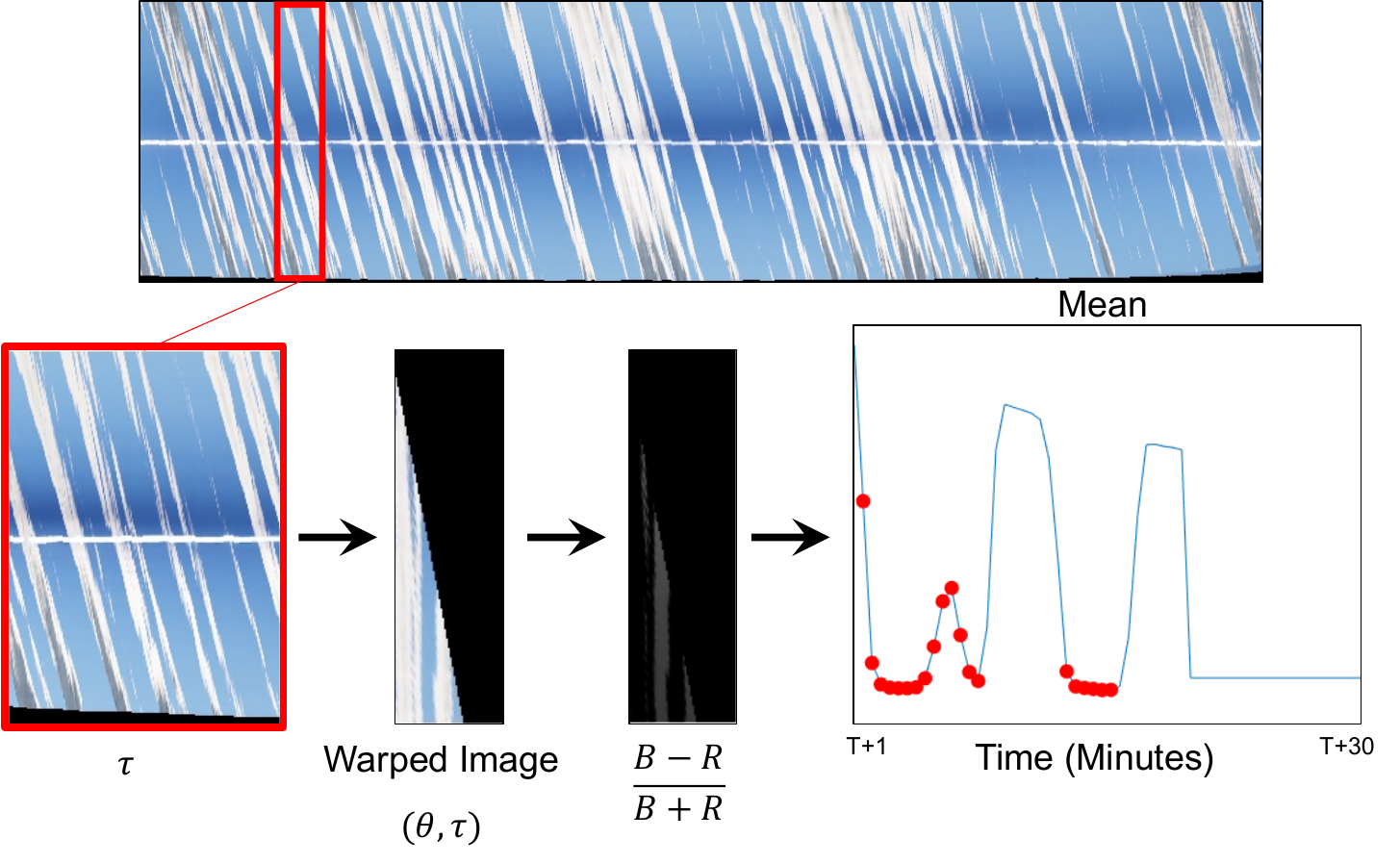}
   \caption{(Top) Space-time image. (Bottom: left-to-right) We visualize the non-learning steps for occlusion prediction. We take some window out of the original space-time image, warp it based on the optimal $\hat{\theta}$, convert it to a new color space based on \cite{HYTA} and plot the mean through time. Red points show ground truth occlusion states. }
    \label{fig:keogram_steps}
%  \label{fig:opticalFlow}
\end{figure}

We take this warped image, and then convert it to a new color space defined as the ratio $(B-R)/(B+R)$ where $B$ and $R$ are the intensities in the blue and red channels of the image, respectively. 
This space allows for a simple contrast-based delineation of cloud versus non-cloud pixels \cite{HYTA}. The mean of this image along the y-axis is computed and used to identify the binary sun occlusion state. As shown in Figure \ref{fig:keogram_steps}, dips in the plotted mean relate to sun occlusion states.

It should be noted that to find the optimal $\hat{\theta}$, we sweep over a range of empirically chosen $\hat{\theta} = \left[60^{\circ}, 85^{\circ} \right]$. Instead of computing the mean, we use the standard deviation and attribute the lowest value of the standard deviation to the optimal $\hat{\theta}$ and warp the image by that value using Eq.\ \eqref{eq:keoEq}.

To obtain metrics in terms of accuracy of the sun occlusion state, we employ the receiver operating characteristic (ROC) curve to obtain the optimal threshold for deciding the occlusion state. We then look at accuracy through time for future time steps using the area under the curve (AUC) which provides an accumulated measurement of performance across all classification thresholds.

\subsection{Learning-Based Occlusion Prediction}
We believe that tasking a learning-based system to understand the dependencies of the space-time image to predict a future sun occlusion state will yield better results than a non-learning approach.

\subsubsection{Neural Occlusion Prediction For Simulated Images} Non-learning based prediction of sun occlusion is limited by the dimensionality of the space-time image. As a result, the time prediction to $T+N$ of a sun occlusion is capped at a certain value of $N$ which is based on $\hat{\theta}$. Using a simple CNN-MLP, we are able to show that a learning-based method can learn the dependency between spatial cloud locations at $\tau$ along with $\hat{\theta}$ to predict the sun occlusions state for a future time instant. Our model is fed the warped image consisting of $\tau$ space-time-slices and produces a $\left(1 \times T+N \right)$ vector which are the binary occlusion state predictions from $T+1$ to $T+N$. Our model is trained end-to-end using binary cross entropy as the loss function. 
%We present comparable results to the non-learning approach for both mirrors in Figure \ref{fig:synth_error_plots}.

\subsubsection{Neural Occlusion Prediction For Real Images}
\label{sec:neural_real}
\begin{figure}[tb]
  \centering
    \includegraphics[width=1\linewidth]{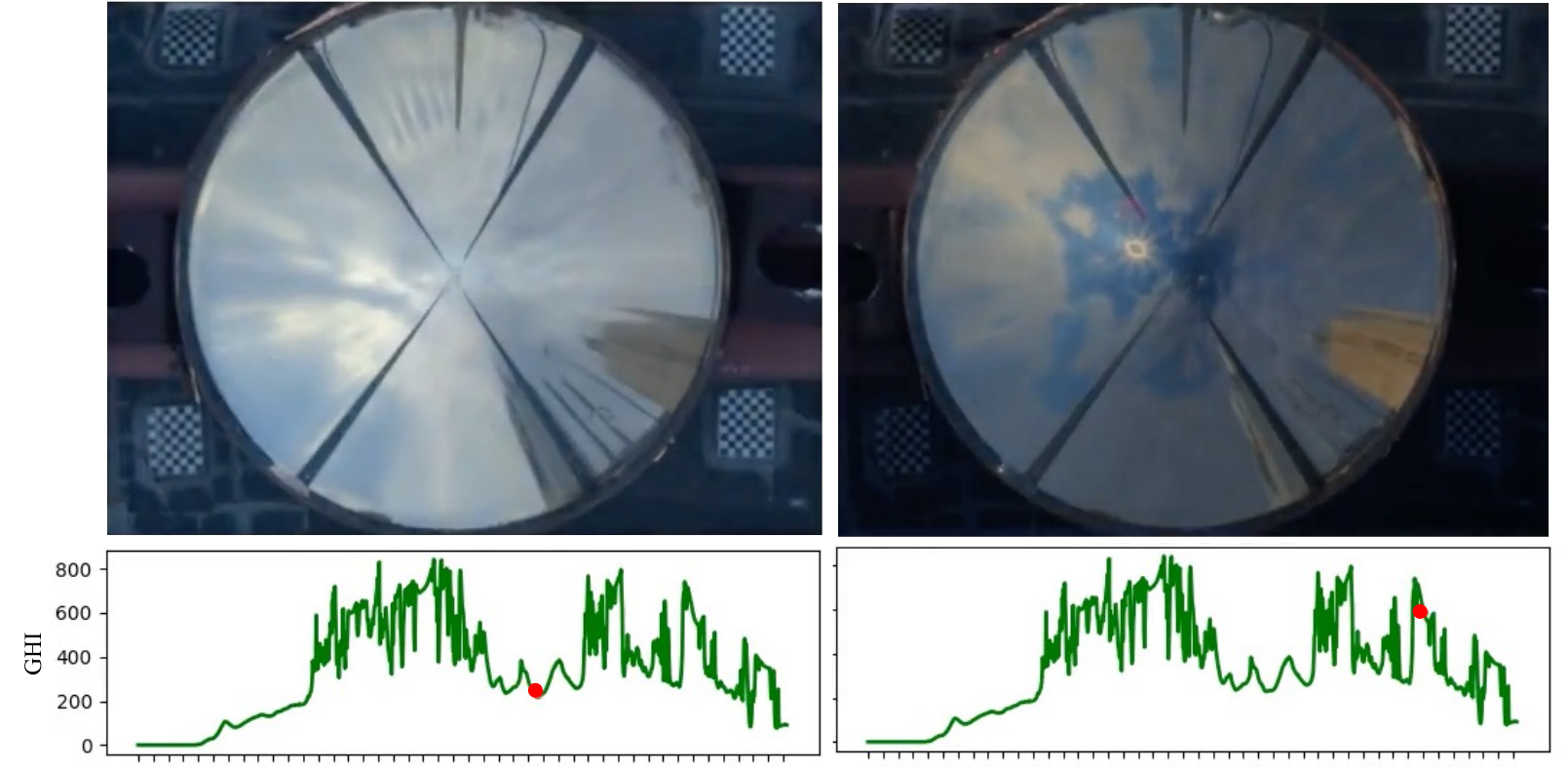}
    \caption{We show sample GHI values based on cloud conditions.}
    \label{fig:architecture}
%  \label{fig:opticalFlow}
\end{figure}

For the real images, we utilize a different approach. Rather than directly predicting the binary occlusion state of the sun, we predict the solar irradiance, in the form of GHI, at a future time instant which directly correlates to sun occlusion. As shown in Figure \ref{fig:architecture}, a decrease in GHI directly correlates to the occlusion of the sun by a cloud and therefore can be used as a prediction method for real images. Overall, GHI is predicted as opposed to the direct sun occlusion state due to the fact that we do not have the ground truth occlusion state for the real images. However, GHI can be deemed an improved occlusion metric because in a binary state, there is no indication of \textit{how} occluded the sun really is. GHI better captures this information and is therefore a more informed prediction value for sun occlusion.
%

%\begin{figure}
%\centering
%            	\includegraphics[, clip, width=0.475\textwidth]{Figures/sample_GHI3.pdf}
%	 \caption{We present sample GHI predictions for the hyperboloidal mirror and the hemispherical mirror. (Left) we input 8 minutes of data into the model to predict 10 minutes of GHI. (Right) We input 30 minutes of data to predict 30 minutes of GHI. For the hyperboloidal mirror, we not only predict the GHI trend close to the ground truth but we are also able to predict the sharp event of when GHI decreases.} 
%	\label{fig:sample_ghi}
%\end{figure}

\par Our learning pipeline for forecasting GHI is 2-fold and very similar to \cite{goswami2024moment} being that we first pre-train our model on masked-input reconstruction followed by fine-tuning for prediction.
Our model architecture employs a transformer encoder \cite{Raffel2019ExploringTL}, and instead of a transformer-based decoder, we use a simple, lightweight reconstruction head for pre-training and a forecasting head for fine-tuning.
Both of these heads are small multi-layer perceptrons (MLPs) consisting of linear and dropout layers.

%%%%GHI + Model architecture %%%%
%%%%%%%%%%%%

\paragraph{Pre-training.} During pre-training, we use information from our space-time slices along with their associated GHI values as input $\{\tau ; \  G_{\tau} \dots G_{T} \}$. The input space-time slice image is sent to a small image encoder consisting of 5 convolution layers where each layer is followed by a ReLU and 2D Maxpool; except for the last layer. This results in a latent embedding ($K$) consisting of 2-channels that encapsulates the information from the space-time slice image.
Concurrently, for the associated GHI values, we employ masked-input reconstruction where, during training, 25\% of the  input is masked-out and replaced  with a learnable mask embedding.
Theoretically, we treat this input GHI as a time series data where information about the current cloud conditions are added via the space-time slice images.
The masked GHI is fed into a patch embedding layer similar to \cite{dosovitskiy2020vit} and the resulting latent embedding ($I$) is concatenated with $K$ and passed into the transformer encoder.
The encoder passes its learned output to the reconstruction head which reconstructs the original masked input.
\par Overall, the goal of this pre-training is to allow the model to learn a representation of the original GHI with cloud information present in the space-time slice image. Figure \ref{fig:grabber} presents a visual of the described steps.
\paragraph{Fine-tuning for forecasting.} The goal of fine-tuning the model to the task of forecasting is to use the pre-trained learned representation of a space-time slice image and its associated GHI value.
During the fine-tuning for forecasting, we replace the reconstruction head with a forecasting head. The forecasting head is again a lightweight MLP consisting of a dropout and linear layer.
Every other weight parameter of the model is frozen during fine-tuning except for the forecasting head.
The model takes the same input of the space-time slice image and its associated GHI values $\{\tau ; \  G_{\tau} \dots G_{T} \}$. However, instead of reconstructing the original input, the model predicts the GHI at future time instances: $[ \widehat{G}_{T+1} \dots \widehat{G}_{T+N}]$.
The goal of this is to limit the amount of trainable parameters for the task of forecasting all while using the high-level features and learned weights from the encoder.
Both pre-trained and fine-tuned models utilize mean-squared error (MSE) as loss functions.

\section{Evaluation}
\label{sec:eval}

\begin{figure}
\centering
\includegraphics[width=0.5\textwidth]{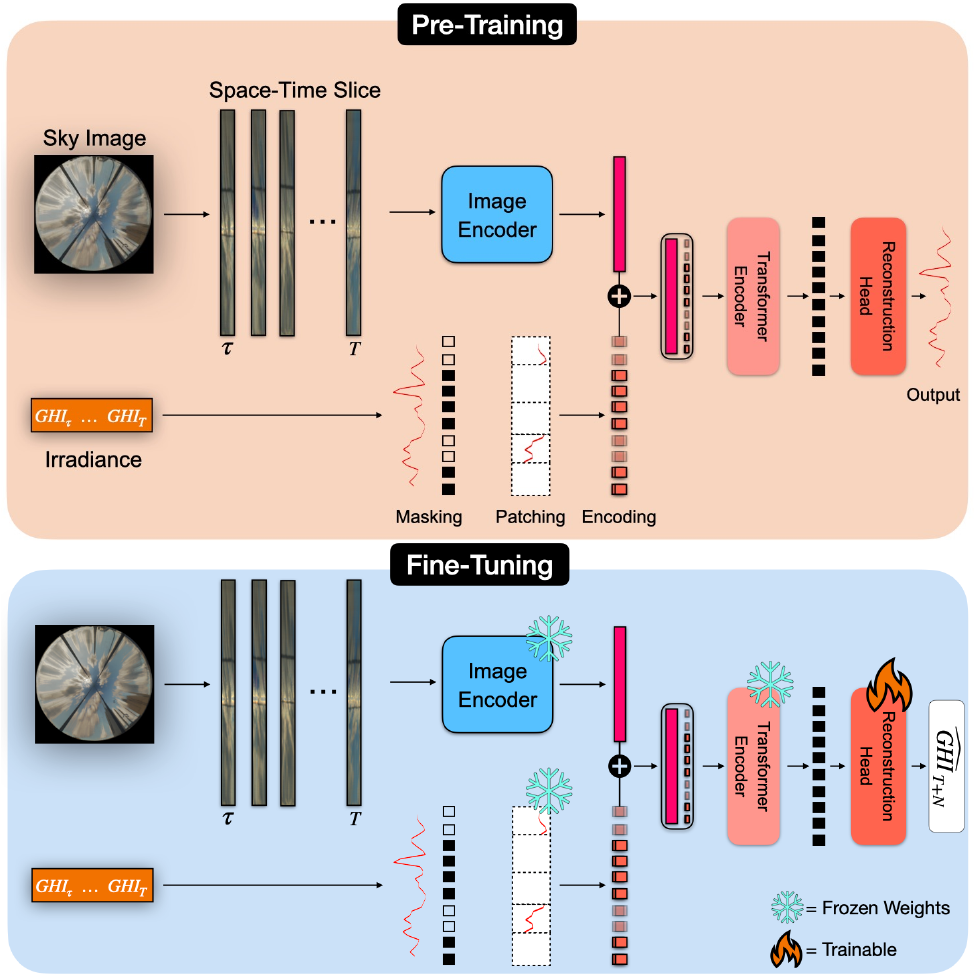}
\vspace{-22pt}
%\caption{Our work presents a computational imaging system based on a catadioptric combination of mirrors and cameras. 
%We initially capture a set of sky images and corresponding irradiance values, extract spatio-temporal slices from each image (see Section \ref{subsec:keogram})
\caption{We forecast future irradiance values via a 2-step end-to-end training pipeline similar to  \cite{goswami2024moment}. We first pre-train on reconstructing the input irradiance followed by fine-tuning for irradiance forecasting. }
%The benefit of our hyperboloidal-based mirror which delivers wide-angle imagery with uniform spatial resolution of the sky over its field of view enables more accurate prediction over a longer time horizon than traditional hemispherical imagers.}
%\vspace{-22pt}
\label{fig:grabber}
\end{figure}

%\begin{figure}[tb]
%            	\includegraphics[width=0.475\textwidth]{Figures/para_gtGHI_vs_Pred_60_60} 
%	\vspace{5mm} \\
%            	\includegraphics[width=0.475\textwidth]{Figures/sphere_gtGHI_vs_Pred_60_60}
%	 \caption{(Top 2 rows-hyperboloidal, Bottom 2 rows-hemispherical). In this figure, we present the predicted GHI on the x-axis plotted against the true GHI values on the y-axis. For intermittent predictions from T+1 to T+30 minutes in the future, a good prediction are linear points that lie on the line. Fed 30 minutes of past data as input, we show that although we obtain similar predictions at T+1 for both mirrors, our mirror outperforms the hemispherical mirror further into the future} 
%	\label{fig:ghi_vs_ghi_plot}
%\end{figure}

%\begin{figure}[tb]
%\centering
%            	\includegraphics[, clip, width=0.475\textwidth]{Figures/sample_GHI2}
%	 \caption{We present sample GHI predictions for the hyperboloidal mirror and the hemispherical mirror. (Left) we input 8 minutes of data into the model to predict 10 minutes of GHI. (Right) We input 30 minutes of data to predict 30 minutes of GHI. For the hyperboloidal mirror, we not only predict the GHI trend close to the ground truth but we are also able to predict the sharp event of when GHI decreases.} 
%	\label{fig:sample_ghi}
%\end{figure}

In this section, we present results from the above algorithms for both the simulated and real data.
\subsection{Simulations}
For experiments on the non-learning based sun occlusion prediction for the simulated data, we use 100 minutes ($\tau=200$) of past data to predict 30 minutes ($N=60$) of sun occlusion state values, in 30 second intervals. Using the algorithms expressed in section \ref{sec:algo}, we achieve promising results.

\begin{figure}[t!] 
\centering
            	\includegraphics[trim={3em 3em 5em 9em}, clip, width=0.375\textwidth]{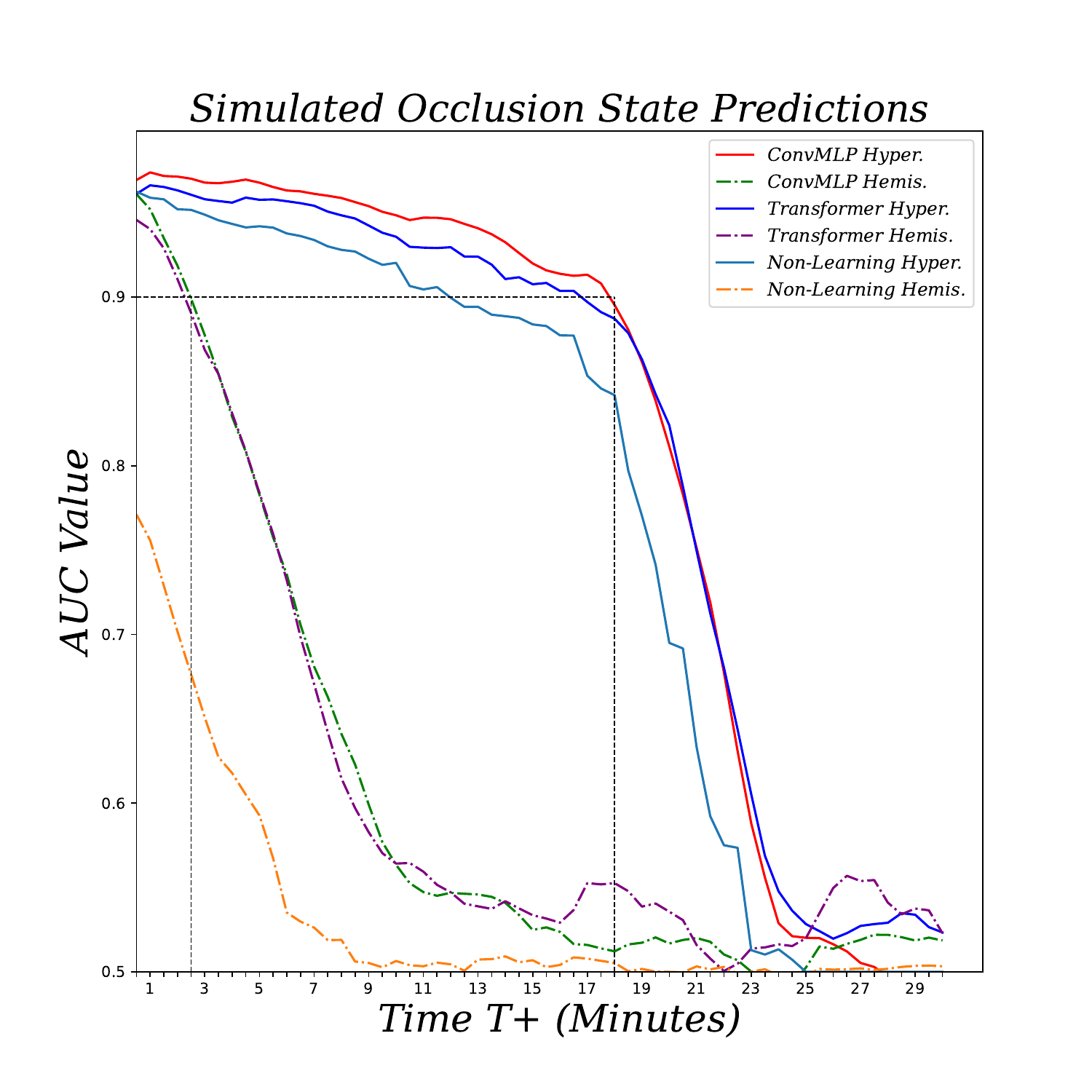}
	 \caption{We compare learning and non-learning based approaches for binary sun occlusion states on the simulated data. Using 90\% as an AUC value for confident predictions, this plot clearly visualizes the stark difference in prediction that we are able to obtain. Looking at the learning-based approaches, for the hemispherical-based mirror, we are limited to a confident forecasting window of around 3 minutes. Compared to hyperboloidal mirror which is able to achieve confident forecasts around 18 minutes into the future. }
	 
	\label{fig:synth_error_plots}
\end{figure}

As seen in Figure \ref{fig:synth_error_plots} we are able to obtain reliable occlusion predictions up to $\approx$ 18 minutes into the future compared to the hemispherical mirror that is only able to maintain solidified predictions to $\approx$ 3 minutes.
Learning-based methods for occlusion prediction provide the best results, providing greater improvement over the back projected method. For the hyperboloidal mirror we benefit with even greater accuracy, longer through time, all while still substantially outperforming predictions obtained using the hemispherical mirror. For comparison, we experimented using a Transformer-based architecture on the simulated data with the same inputs. Simulated results clearly show the benefit of using a hyperboloidal mirror setup coupled with a learning-based system for long-term prediction of sun occlusion by a cloud. We now present real-world results and show the benefits of using our system.

\begin{table*}[!t]
\caption{We present comparison results on various models trained and tested on the same datasets. The Transformer model is the proposed architecture in Section \ref{sec:neural_real}. "Combined" concatenates the space-time-slice image from the hyperboloidal and the hemispherical mirror in the channel dimension. The Seq2Seq model exploits recurrence in the form of gated recurrent units (GRUs) to model the sequential aspect of the space-time-slice images along with the associated GHI. The values are nRMSE between ground truth and predicted GHI values at future time instants.}
\label{model_zoo}
\ra{1.3}
\centering
\begin{tabular}{@{}cccccccc@{}} \toprule
Model & Data & {1 min} & {5  min}& {10 min} & {15  min} & {20 min}& {30 min}\\
\midrule
 Persistence & GHI alone & 0.191&0.267&0.300&0.323&0.340&0.336 \\
 Seq2Seq & GHI alone & \textbf{0.169} & 0.233 & 0.265 & 0.287 & 0.304 & 0.307 \\
 \specialrule{.005em}{0em}{0em} 
    Seq2Seq  & Hemispherical + GHI  & 0.185&0.227&0.239&0.255&0.271&0.270 \\
    Transformer  &Hemispherical + GHI & 0.192&0.236&0.250&0.258&0.273&0.284 \\
    Transformer  & Hemispherical Warped + GHI& 0.190&0.236&0.253&0.294&0.310&0.341 \\
     \specialrule{.005em}{0em}{0em} 
         Seq2Seq &Hyperboloidal + GHI & {0.179}&0.227&0.248&0.260&0.277&0.277 \\
    Transformer  & Hyperboloidal + GHI & 0.191&0.227&\textbf{0.230}&\textbf{0.246}&0.266&0.289 \\
    Transformer  &Hyperboloidal Warped + GHI & 0.189&0.233&0.266&0.294&0.303&0.319 \\
         \specialrule{.005em}{0em}{0em} 
    Transformer & Combined + GHI & 0.183&\textbf{0.223}&0.239&0.252&\textbf{0.259}&\textbf{0.268} \\
\bottomrule
\end{tabular}
%}
\end{table*}

\subsection{Real Data}
For experiments using real data, we pass 30 minutes ($\tau=60$) of data to predict 30 minutes ($N=60$) of GHI, in 30 second intervals. Although as not ideal as in the simulated case, results from the real case still achieve GHI prediction performance over the hemispherical mirror. For our accuracy metric, we use the normalized root mean-squared error (nRMSE):
\begin{equation}
	RMSE = \sqrt{\dfrac{\sum_{n=1}^{N}  \left( \hat{y}_{n} - y_{n} \right)^{2}}{N}}.
\end{equation}
\begin{equation}
	nRMSE = RMSE / \sqrt{\dfrac{\sum_{n=1}^{N} y_{n} ^{2}}{N}}.
\end{equation}
where $\hat{y}_{n}$, $y_{n}$ is the predicted GHI and true GHI, respectfully. A lower value equates to a more accurate prediction.

\begin{figure}[t!]
\centering
            	\includegraphics[trim={2em 3em 5em 9em}, clip, width=0.375\textwidth]{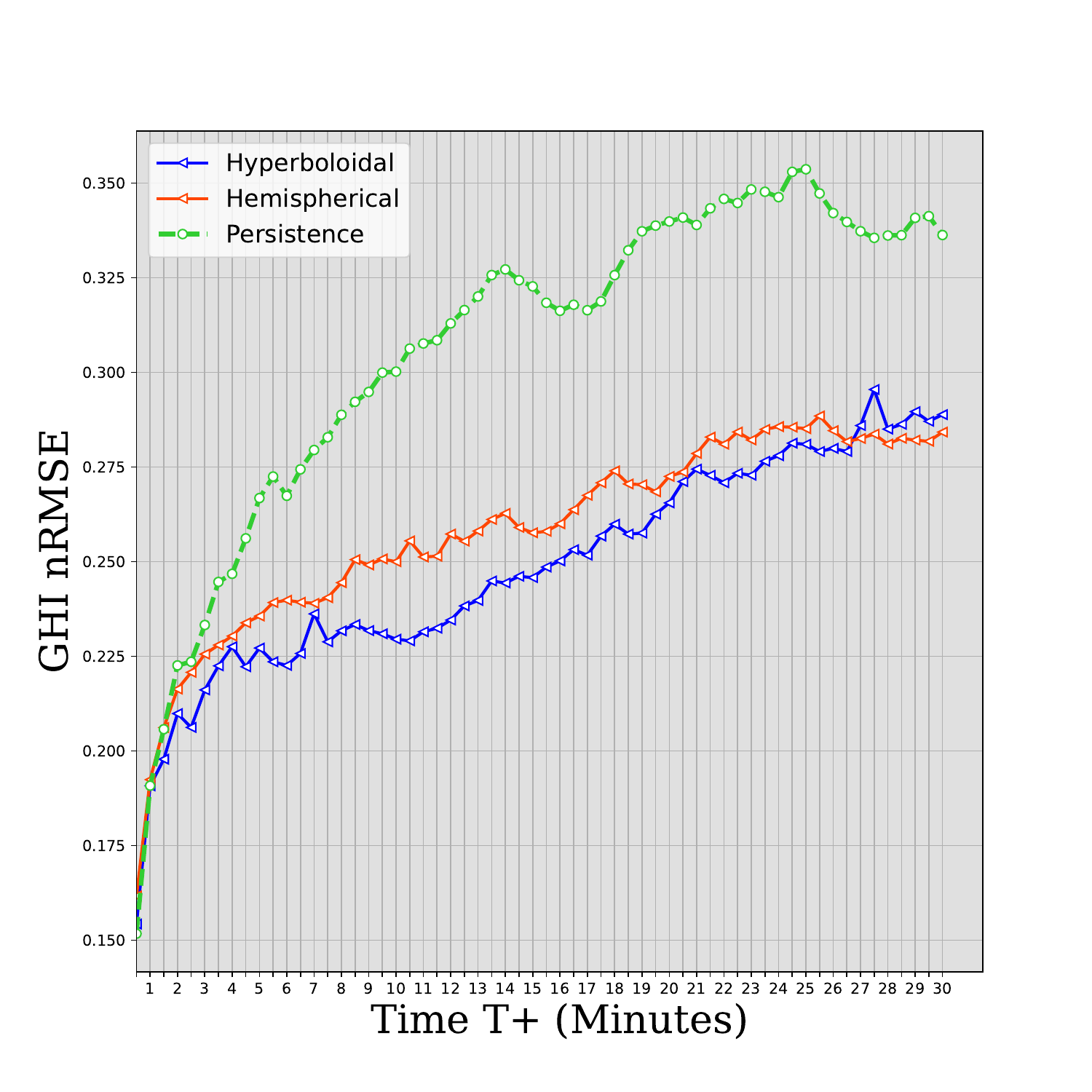}
	 \caption{This figure shows the outperformance of the hyperboloidal mirror. Our mirror is able to achieve lower forecasting error longer into the future compared to the traditional hemispherical mirror. Both models are trained using the model described in section \ref{sec:neural_real} which is fed 30 minutes of past data to predict 30 minutes of GHI. As a baseline comparison, we compare both imaging setups to the persistence model which states that the irradiance value will remain unchanged over the forecasting horizon ($\widehat{GHI}_{t+ \Delta t} = GHI_{t}$). Our model still outperforms the persistence model longer into the future.}
	\label{fig:real_error_plot}
\end{figure}

We present results of prediction values in Figure \ref{fig:real_error_plot} which shows that we are able to predict GHI longer into the future with lower error compared to the hemispherical mirror.
As a baseline comparison, we also compare both imaging setups to the persistence model which states that the irradiance value will remain unchanged over the forecasting horizon ($\widehat{GHI}_{t+ \Delta t} = GHI_{t}$).

\subsection{Ablation}
In Table \ref{model_zoo}, we present an ablation study that shows how different model architectures and different data sources affect the prediction results. 
For the models, we used two models: \textit{Transformer:} the proposed architecture in Section \ref{sec:neural_real}, and \textit{Seq2Seq:} a RNN architecture comprised of  gated recurrent units (GRUs)  that model the sequential aspect of the space-time-slice images along with the associated GHI.
For the data sources, we have access to the GHI measurements from the pyranometer, and the imagery from the hemispherical and hyperboloidal cameras.
We trained models on various combinations as shown in the Table.
Finally, as a reference, we report results on a commonly-used baseline  called `persistence' where the latest GHI value is used as the prediction for all future time instants.
The results in Table \ref{model_zoo} suggest that, except for the shortest time horizon, models trained with hyperboloidal data outperforms the rest.
For the shortest time horizon of 1 minute, GHI alone model works the best; given that all models have access to the GHI data, we believe this is likely a manifestation of the small amount of data compared to the variability of the sky imagery.
%
%
%We wanted to \textit{feed} the model the prediction information and exploit the benefits of our hyperboloidal setup. 
%
%"Combined" concatenates the space-time-slice image from the hyperboloidal and the hemispherical mirror in the channel dimension. 
%
%The hyperboloidal imager captures long-term predictions compared to the hemispherical imager.
%
%What information can be learned if we combined both images in an attempt to improve prediction results?
%
%That is what this input variation attempted to understand.
%%
%The Seq2Seq model exploits recurrence in the form of gated recurrent units (GRUs) to model the sequential aspect of the space-time-slice images along with the associated GHI.
%
%When comparing these models using nRMSE as a prediction metric, no architecture clearly stands out. 
%
%This could be due to each model learning and focusing on different aspects of the input, leading to variations of the predicted outputs.

\section{Discussions}
\label{sec:discuss}

This paper argues for a novel system that brings core computational imaging techniques to a compelling problem in renewable energy.
Our work provides a pathway to improve the time horizon over which we can reliably forecast solar irradiance; specifically, over conventional wide FoV systems, we can improve predictions from minutes to tens of minutes.
We expect such a prediction framework to be of wide interest in the solar photovoltaics community, where resource allocation and energy dispatch is often done in the absence of such predictive analytics.
Finally, on a broader scale, we hope the techniques suggested in this paper continue to incite the interest in applications that lie at the intersection of imaging and climate change. 

\paragraph{Beyond hyperboloidal mirrors.}
Our choice of a hyperboloidal shaped mirror provides a single viewpoint system while preserving uniform resolution over the sky plane.
A natural  question to consider is whether either of these assumptions are need for solar forecasting.
For example, we could consider alternate designs that place a larger emphasis on resolution at the horizon at the cost of even reduced resolution at the zenith.
Similarly, there is no critical reason why designs like conic mirrors that do not offer a single viewpoint \cite{710698} are not applicable here.
Effectively, we could enable a framework where the solar forecasting  guides the design or optimization of the mirror profile in an end-to-end manner.
%clouds that are further out at the horizon better determine what its evolution will lead to right above you,
%we should put a larger emphasis on that location. 
%Future work could develop imaging systems that focus on clouds at the horizon in an effort to increase forecasting times even greater.

\paragraph{Cloud formation and disappearance.}
One of the factors that we fail to consider in this paper is that clouds form and disappear based on changes in humidity, temperature, and pressure.
This violates the slicing model used in this work, in part because clouds can appear in the middle of the field of view, or disappear as it traverses the field.
In our testbed, this happens frequently at a particular spot that is over a water body, a few miles from  the deployed system.
The spatial consistency of the cloud formation suggests that statistical models that have a better understanding of how clouds form and terminate, augmented with other sources of data such as weather, humidity and the geographic layout of the surrounding regions, might have a better chance in handling the effects.

\paragraph{Self-occlusion by clouds.}
Another factor that we fail to consider is that clouds have vertical extent, and hence a cloud closer to the camera may block one that is further away.
This shows itself in the form of radial streaks in Figures \ref{fig:synthGallery} and \ref{fig:realImgGallery}.
This is a hard problem to resolve in the absence of additional view points.
It is likely that a multi-camera version of our system with a baseline in kilometers will be able to reason such occlusions and handle them effectively.

\paragraph{Incorporating other data sources.}
The techniques proposed in this paper will also benefit from other richer sources of data such as satellite imagery, weather prediction, and humidity measurements.
Such sources of data are often publicly available; however, each of them have unique features that need to be accounted for. 
For example, satellite data that provides very large spatial extent, has very poor temporal resolution, often in minutes, if not hours.
Further, the ground spatial resolution of such data is in meters, which may not be sufficient for the kind of prediction we envision.
Wind velocity, which is often available from weather data, is something we can benefit from.
However, such measurements are often made at ground level, and at very sparse locations, which limits their utility, since atmospheric wind velocities differ from ground measurements.
Yet, the role that temperature, humidity, and, more broadly, the weather play in forecasting cannot be denied. 
Translating such models to near-future time horizons and higher precision that is demanded for solar forecasting is an interesting approach for subsequent research.

%\paragraph{Prediction beyond a single location}

% Any acknowledgments to only be included in camera ready
\ifpeerreview \else
\section*{Acknowledgments}
This work was supported, in part, by the NSF EAGER award 2235063. Julian acknowledges support from a GEM Fellowship and the Fritsch Family Fellowship. Lee was supported in part by a TCS Presidential Fellowship.
\fi

\bibliographystyle{IEEEtran}
\bibliography{skycam}

\ifpeerreview \else
%%%% For the camera ready version, please fill out this
%%%% biography. Your camera ready should be within a 12 page limit
%%%% including acknowledgments, references and biography.

% \includegraphics command within an optional argument. (You could
% create your own custom macro containing the \includegraphics command
% to make things simpler here.)
% \begin{IEEEbiography}[{\includegraphics[width=1in,height=1.25in,clip,keepaspectratio]{mshell}}]{Michael Shell}

%\vfill\eject
\vspace{-22pt}
\begin{IEEEbiography}[{\includegraphics[width=1in,height=1.25in,clip,keepaspectratio]{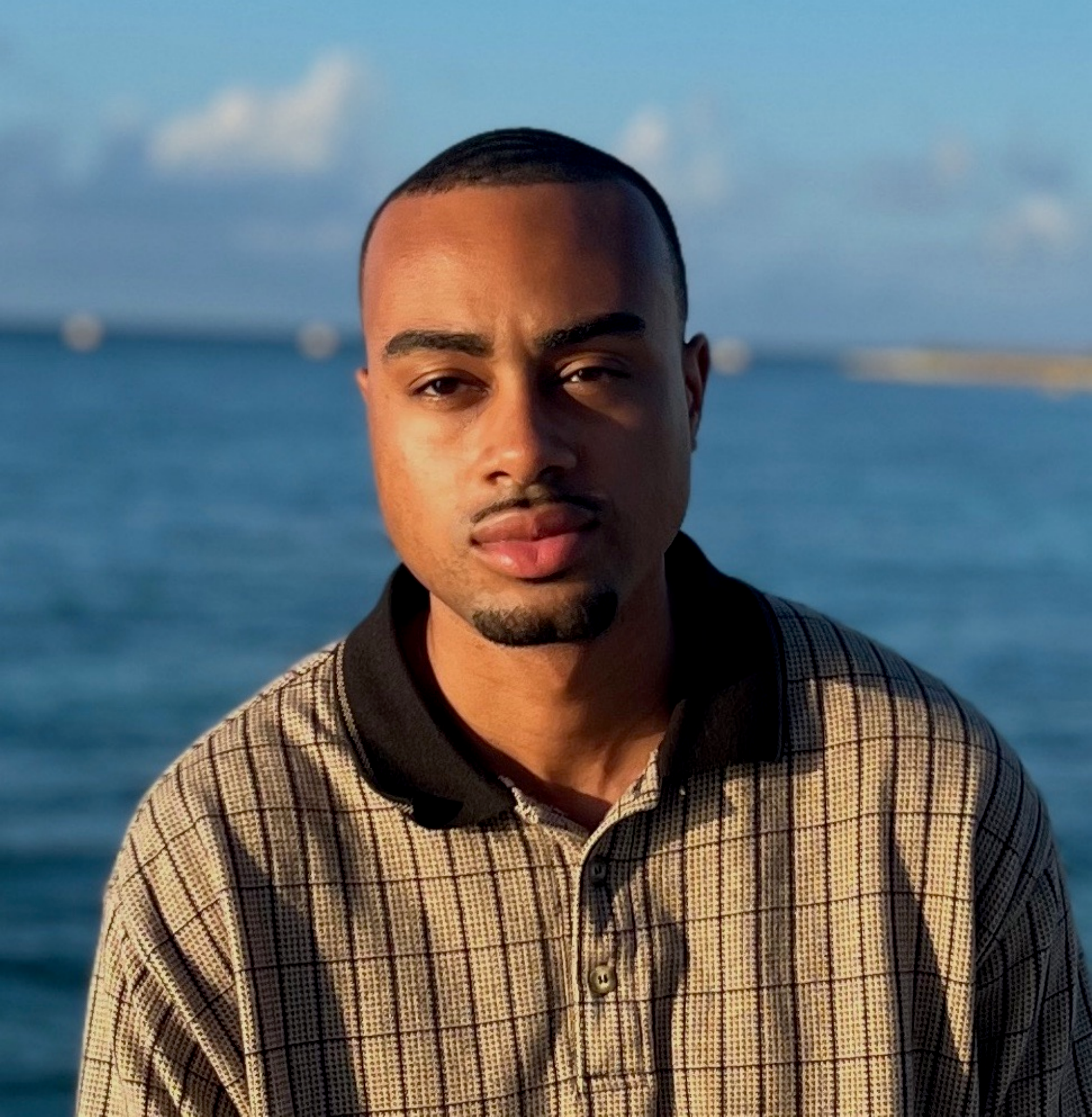}}]{Leron Julian} received his B.S. in Computer Science at Morehouse College in 2019 and is a  Ph.D candidate at Carnegie Mellon University. His research aims to increase the efficiency of solar power by leveraging computer vision, computational photography, and learning to capture and forecast the relationship between atmospheric conditions and received solar irradiance at the ground. He is also the recipient of a GEM fellowship and the Fritsch Family fellowship.
\end{IEEEbiography}
\begin{IEEEbiography}[{\includegraphics[width=1in,height=1.25in,clip,keepaspectratio]{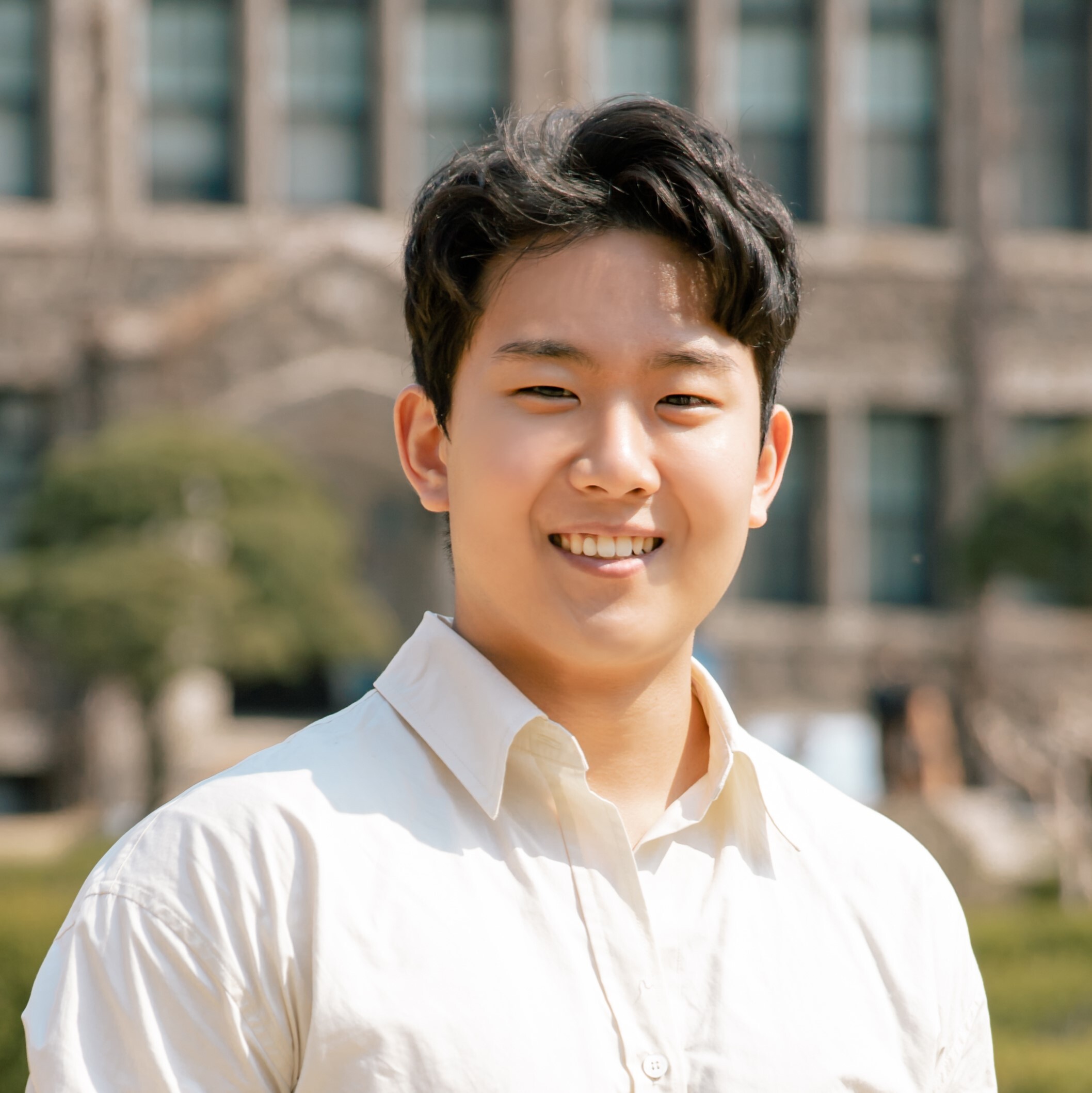}}]{Haejoon Lee} received his B.S. in Electrical and Electronic Engineering from Yonsei University in 2021 and is currently a Ph.D. student at Carnegie Mellon University. His research focuses on computer vision and computational imaging, with applications ranging from solar irradiance prediction to material classification. He is the recipient of the Presidential Fellowship funded by Tata Consulting Services in 2024 and the Korean Government Scholarship for Study Overseas in 2023.
\end{IEEEbiography}
\begin{IEEEbiography}
[{\includegraphics[width=1in,height=1.25in,clip,keepaspectratio]{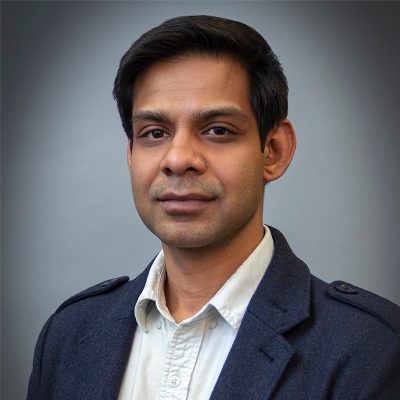}}]{Soummya Kar} is the Buhl Professor of Electrical and Computer Engineering at Carnegie Mellon University. He received a B.Tech. in electronics and electrical communication engineering from the Indian Institute of Technology, Kharagpur, India, in May 2005 and a Ph.D. in electrical and computer engineering from Carnegie Mellon University, Pittsburgh, PA, in 2010. 
%From June 2010 to May 2011, he was with the Electrical Engineering Department, Princeton University, Princeton, NJ, USA, as a Postdoctoral Research Associate. 
His research interests include decision-making in large-scale networked systems, stochastic systems, multi-agent systems and data science, with applications in cyber-physical systems and smart energy systems. He is a Fellow of the IEEE.
\end{IEEEbiography}
\begin{IEEEbiography}
[{\includegraphics[width=1in,height=1.25in,clip,keepaspectratio]{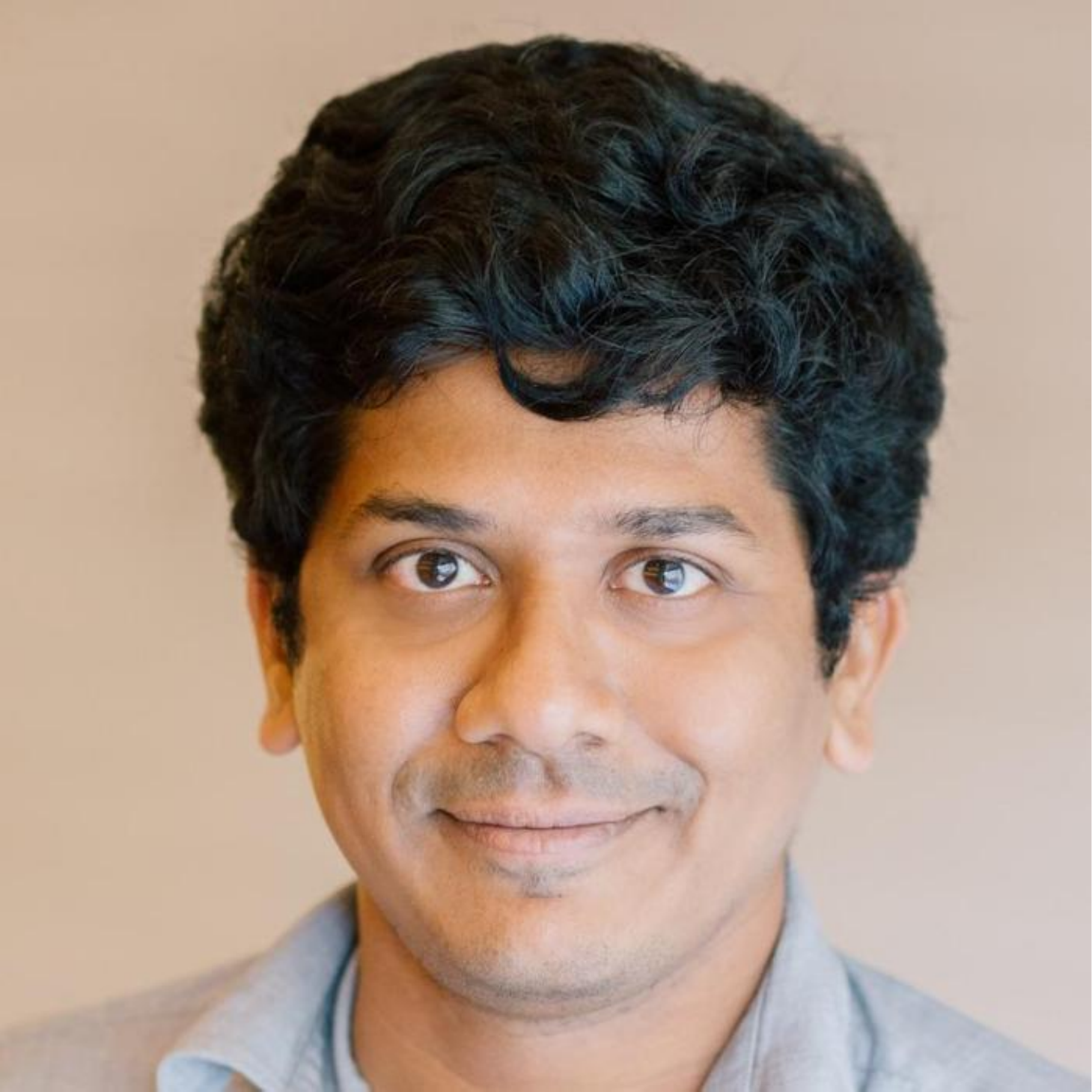}}]{Aswin C. Sankaranarayanan}
is a professor in the ECE department at CMU, where he is the PI of the Image Science Lab. His research interests are broadly in compressive sensing, computational photography, signal processing and machine vision. His doctoral research was in the University of Maryland where his dissertation won the distinguished dissertation award from the ECE department in 2009. Aswin is the recipient of best paper awards at SIGGRAPH 2023, CVPR 2019 and ICCP 2021 \& 2022, the CIT Dean’s Early Career Fellowship, the Spira Teaching award, the NSF CAREERaward, the Eta Kappa Nu (CMU Chapter) Excellence in Teaching award, and the Herschel Rich Invention award from Rice University.
\end{IEEEbiography}

%\newpage

%\begin{IEEEbiographynophoto}{Leron Julian}
%\acs{add bios}

%\vfill

\fi

\end{document}